\pdfoutput=1

\documentclass[11pt]{article}

\usepackage[preprint]{acl}

\usepackage{times}
\usepackage{latexsym}
\usepackage{pifont}

\newcommand{\cmark}{\ding{51}} 
\newcommand{\xmark}{\ding{55}} 
\usepackage[T1]{fontenc}

\usepackage[utf8]{inputenc}

\usepackage{microtype}

\usepackage{inconsolata}

\usepackage{graphicx}
\usepackage{booktabs}
\usepackage{siunitx}
\usepackage{array}
\usepackage{colortbl}
\usepackage{xcolor}
\usepackage{booktabs}
\usepackage{threeparttable}
\usepackage{tabularx}
\usepackage[breakable]{tcolorbox}
\tcbuselibrary{skins,breakable}  
\usepackage{listings} 
\tcbset{
  remarkbox/.style={
    enhanced,
    breakable,
  }
}

\usepackage{pdflscape}
\usepackage{longtable}
\usepackage{rotating}

\definecolor{qwencolor}{RGB}{230,240,255}
\definecolor{llamacolor}{RGB}{255,240,230}
\definecolor{deepseekcolor}{RGB}{230,255,230}
\definecolor{phicolor}{RGB}{240,230,255}
\definecolor{gemmacolor}{RGB}{255,245,230}
\definecolor{mistralcolor}{RGB}{230,250,255}    
\definecolor{ayacolor}{RGB}{255,230,245}
\definecolor{interncolor}{RGB}{245,245,230}
\definecolor{gptcolor}{RGB}{240,240,255}

\definecolor{mistraltextcolor}{RGB}{178,157,196}
\definecolor{phitextcolor}{RGB}{152,183,176}
\definecolor{qwentextcolor}{RGB}{189,148,135}
\definecolor{llamatextcolor}{RGB}{139,134,128}
\definecolor{deepseektextcolor}{RGB}{115,128,125}
\definecolor{gemmatextcolor}{RGB}{90,121,122}
\definecolor{gpttextcolor}{RGB}{57,94,95}
\definecolor{internlmtextcolor}{RGB}{24,66,67}
\definecolor{ayatextcolor}{RGB}{109,109,0}   
\definecolor{claudecolor}{RGB}{220,230,245}

\definecolor{westeuropean}{RGB}{128,155,206}    
\definecolor{indoaryan}{RGB}{149,184,209}       
\definecolor{eastasian}{RGB}{184,224,212}       
\definecolor{african}{RGB}{214,234,223}         
\definecolor{overall}{RGB}{229,212,239}         
\definecolor{other}{RGB}{255,214,165}

\title{\textit{MMLU-ProX}: A Multilingual Benchmark for \\Advanced Large Language Model Evaluation}

\author{\textbf{Weihao Xuan}$^{1*}$, \textbf{Rui Yang}$^{2}$, \textbf{Heli Qi}$^{3}$, \textbf{Qingcheng Zeng}$^{4}$, \textbf{Yunze Xiao}$^{5}$, \textbf{Aosong Feng}$^{6}$, \\
\textbf{Dairui Liu}$^{7}$, \textbf{Yun Xing}$^{8}$, \textbf{Junjue Wang}$^{1}$, \textbf{Fan Gao}$^{1}$, \textbf{Jinghui Lu}$^{9}$, \textbf{Yuang Jiang}$^{9}$, \\
\textbf{Huitao Li}$^{2}$, \textbf{Xin Li}$^{2}$, \textbf{Kunyu Yu}$^{2}$, \textbf{Ruihai Dong}$^{7}$, \textbf{Shangding Gu}$^{10}$, \textbf{Yuekang Li}$^{11}$, \\
\textbf{Xiaofei Xie}$^{12}$, \textbf{Felix Juefei-Xu}$^{13}$, \textbf{Foutse Khomh}$^{14}$, \textbf{Osamu Yoshie}$^{3}$, \textbf{Qingyu Chen}$^{6}$, \\
\textbf{Douglas Teodoro}$^{15}$, \textbf{Nan Liu}$^{2}$, \textbf{Randy Goebel}$^{16}$, \textbf{Lei Ma}$^{1}$, \textbf{Edison Marrese-Taylor}$^{1}$, \\
\textbf{Shijian Lu}$^{8}$, \textbf{Yusuke Iwasawa}$^{1}$, \textbf{Yutaka Matsuo}$^{1}$, \textbf{Irene Li}$^{1*}$ \\
\\
$^{1}$The University of Tokyo, $^{2}$Duke-NUS Medical School, $^{3}$Waseda University, \\
$^{4}$Northwestern University, $^{5}$Carnegie Mellon University, $^{6}$Yale University, \\
$^{7}$University College Dublin, $^{8}$Nanyang Technological University, $^{9}$Smartor LLC, \\
$^{10}$University of California, Berkeley, $^{11}$University of New South Wales, \\
$^{12}$Singapore Management University, $^{13}$New York University, $^{14}$Polytechnique Montréal, \\
$^{15}$University of Geneva, $^{16}$University of Alberta \\
\texttt{weihaoxuan@g.ecc.u-tokyo.ac.jp}, \texttt{irene.li@weblab.t.u-tokyo.ac.jp} \\
\color{magenta}\url{https://mmluprox.github.io/}
}

\begin{document}
\maketitle
\begin{abstract}
Existing large language model (LLM) evaluation benchmarks primarily focus on English, while current multilingual tasks lack parallel questions that specifically assess cross-linguistic reasoning abilities. This dual limitation makes it challenging to comprehensively assess LLMs' performance in the multilingual setting. To fill this gap, we introduce \textit{\textbf{MMLU-ProX}}, a comprehensive benchmark covering 29 languages, built on an English benchmark. Each language version consists of 11,829 identical questions, enabling direct cross-linguistic comparisons. Additionally, to meet efficient evaluation needs, we provide a lite version containing 658 questions per language. To ensure the high quality of \textit{\textbf{MMLU-ProX}}, we employ a rigorous development process that involves multiple powerful LLMs for translation, followed by expert review to ensure accurate expression, consistent terminology, and cultural relevance. Building on this, we systematically evaluate 36 state-of-the-art LLMs, including reasoning-enhanced and multilingual-optimized LLMs. The results reveal significant disparities in the multilingual capabilities of LLMs: While they perform well in high-resource languages, their performance declines markedly in low-resource languages, with gaps of up to 24.3\%. Through \textit{\textbf{MMLU-ProX}}, we aim to advance the development of more inclusive AI systems and promote equitable access to technology across global contexts.

\end{abstract}

\section{Introduction}
The rapid development of large language models (LLMs) has significantly reshaped the field of natural language processing (NLP), with an increasing shift from predominantly English-centric systems towards multilingual understanding~\cite{yang2025qwen3, grattafiori2024llama, aryabumi2024aya}. As LLMs become more prevalent in global applications, the need for comprehensive multilingual evaluations becomes paramount. An effective multilingual evaluation ensures the global accessibility of LLMs, particularly benefiting users of diverse linguistic and cultural backgrounds~\cite{Poppi2024TowardsUT,Bang2023AMM}.

\begin{figure}[t]
    \centering
    \includegraphics[width=0.9\linewidth]{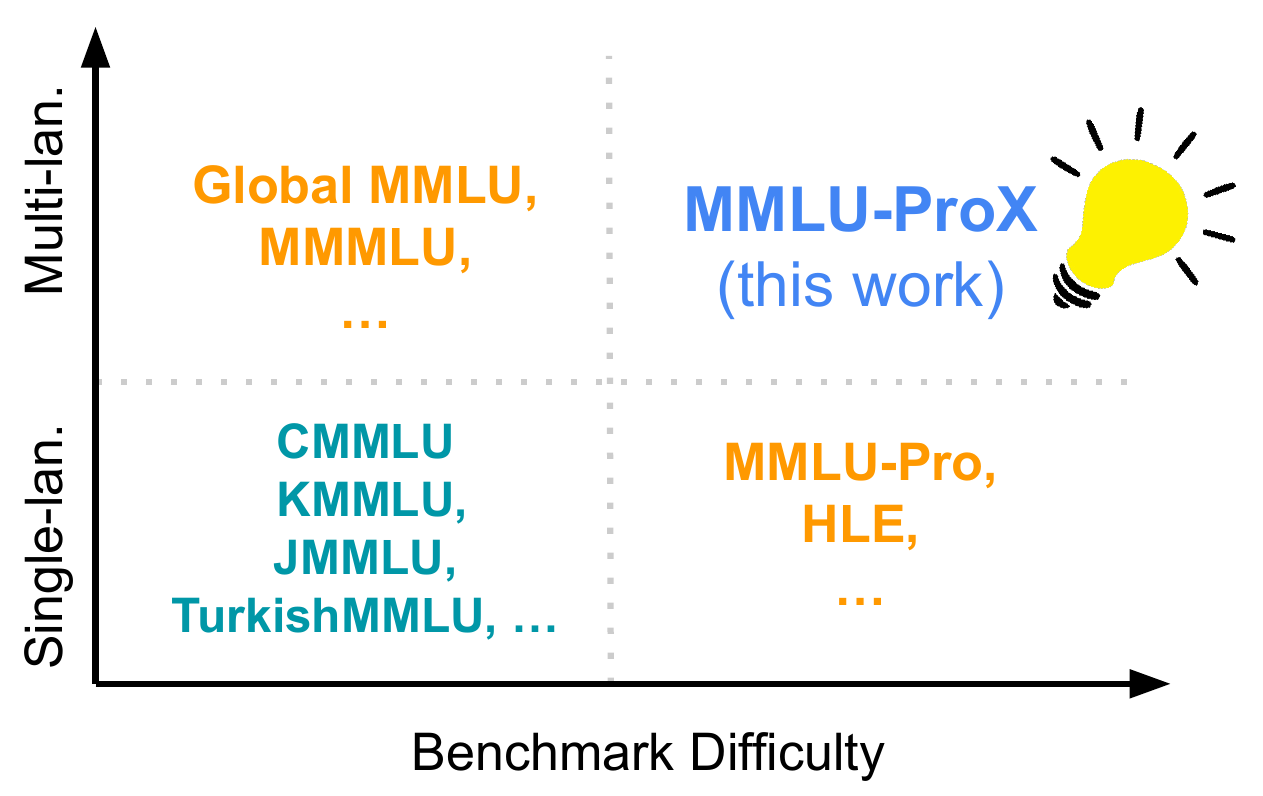}
    \caption{Selected existing benchmarks for multilingual LLMs evaluation on benchmark difficulty and number of languages.}
    \label{fig:teaser}
    \vspace{-5mm}
\end{figure}

The multilingual evaluation of LLMs faces two primary challenges, as illustrated in Figure \ref{fig:teaser}. First, existing benchmarks are constrained by limitations in both language coverage or translation quality. Although monolingual benchmarks such as TurkishMMLU~\cite{yuksel-etal-2024-turkishmmlu}, KMMLU~\cite{son-etal-2025-kmmlu}, and JMMLU~\cite{jmmlu} offer rigorous evaluation within their respective languages, they provide limited insight for comprehensive multilingual evaluation. Broader initiatives such as Global MMLU~\cite{singh2025globalmmluunderstandingaddressing} extend coverage to 42 languages, distinguishing between culture-sensitive and culture-agnostic questions. However, the heterogeneous translation approaches pose significant challenges. The combination of professional translators, community volunteers, and Google Translate introduces quality variations that are difficult to quantify. These inconsistencies in translation quality impede objective comparison of model reasoning across languages and hinder precise diagnosis of low-resource language deficiencies. The second challenge pertains to the difficulty of the evaluation. The evolution from MMLU~\cite{hendrycks2021measuringmassivemultitasklanguage} to MMLU-Pro~\cite{wang2024mmluprorobustchallengingmultitask}, and Humanity's Last Exam (HLE) ~\cite{phan2025humanity} in English benchmarking reflects the rapidly advanced reasoning capabilities of LLMs. Among those, MMLU-Pro enhances its predecessor through more complex reasoning questions, expanded answer choices, and reduction of data set noise, offering greater discriminative power. This progression underscores the pressing need for equally challenging multilingual benchmarks that can effectively evaluate sophisticated reasoning capabilities across languages.

To address these challenges, we introduce MMLU-ProX, a novel multilingual benchmark that builds upon the challenging, reasoning-focused design of MMLU-Pro while extending its coverage to 29 typologically diverse languages. The resulting benchmark contains 11,829 questions per language in its full version, with a lite version of 658 questions available for efficient evaluation. To ensure linguistic accuracy and terminological consistency across languages, we develop a semi-automated translation framework that combines state-of-the-art (SOTA) LLMs with expert verification. This approach effectively mitigates the quality variations inherent in heterogeneous translation methods and maintains the discriminative power of MMLU-ProX in the multilingual setting.

Our primary contributions include: 1) We introduce MMLU-ProX, a multilingual benchmark for massive multitask language understanding with enhanced reasoning-focused questions across 29 languages. It enables comprehensive evaluation of LLMs' cross-lingual reasoning abilities and lays a foundation for the development of more inclusive LLMs in the future. Additionally, we engage over 30 experts to verify the data quality, with a total labor effort exceeding 400 hours. 2) We conduct systematic evaluations on MMLU-ProX and its lite version using both 0-shot and 5-shot chain-of-thought (CoT)~\cite{Wei2022ChainOT} prompting across 36 latest LLMs, covering both open-weight LLMs ranging from 3.8B to 671B parameters, as well as proprietary LLMs. 3) We analyze the reasoning capabilities of LLMs in the multilingual setting, revealing significant performance disparities across languages. This analysis underscores the limitations of current LLMs in global contexts, further highlighting the need to enhance global accessibility and advance fairness evaluations.

\section{Related Work}

\begin{table*}[ht]
\centering
\resizebox{\textwidth}{!}{%
\begin{tabular}{lclcccl}
\toprule
\textbf{Dataset} & \textbf{Languages} & \textbf{Evaluation Modality} & \textbf{CoT} & \textbf{Parallel Data} & \textbf{Subjects} & \textbf{Questions} \\
\midrule
MMLU \cite{hendrycks2021mmlu} & 1 & Multiple-choice (4) & \xmark & \xmark & 57 & 15908\\ 
TurkishMMLU* \cite{yuksel-etal-2024-turkishmmlu}& 1 & Multiple-choice (4) & \xmark & \xmark & 9& 10032\\ 
KMMLU \cite{son-etal-2025-kmmlu}& 1& Multiple-choice (4) & \xmark & \xmark & 45&35030\\
XCOPA \cite{ponti-etal-2020-xcopa} & 11 & Binary choice & \xmark & \xmark & 1 & 5500\\
Global-MMLU \cite{singh2025globalmmluunderstandingaddressing} & 42 & Multiple-choice (4) & \xmark & \cmark & 57 & $\approx 600$k\\
MMMLU\footnote{\url{https://huggingface.co/datasets/openai/MMMLU}} & 14 & Multiple-choice (4) & \xmark & \cmark & 57 & $\approx$ 197k \\
Humanitiy's Last Exam\cite{phan2025humanity} & 1 & Multiple-choice \& exact match & \xmark & \xmark & 2 & $\approx$ 5000\\
MMLU-Pro \cite{wang2024mmluprorobustchallengingmultitask} & 1 & Multiple-choice (10) & \cmark & \xmark & 57 & $\approx$ 12k\\
\rowcolor{overall}
MMLU-ProX (this work) & 29 & \textbf{Multiple-choice (10)} & \textbf{\cmark} & \textbf{\cmark} & 57 & $\approx$ 342.2k \\
\bottomrule
\end{tabular}%
}
\caption{
Comparison of multilingual benchmarks with ticks (\cmark) and crosses (\xmark) indicating presence or absence of CoT and Parallel Data. *We acknowledge other MMLU datasets for various languages and randomly select TurkishMMLU as a representative example.}
\label{tab:compare}
\vspace{-3mm}
\end{table*}

\noindent \textbf{Multilingual Large Language Models.}
The field of NLP has been profoundly transformed by multilingual LLMs, which have evolved beyond the initial English-centric paradigm to address the linguistic diversity of our world with over 7,000 languages spoken globally~\cite{etxaniz-etal-2024-multilingual}. Modern LLMs are sophisticated systems built upon advanced neural architectures such as the Transformer, designed to process, comprehend, and generate text across numerous languages. Recent LLMs such as Claude 3 series~\cite{anthropic2025claude}, GPT-4~\cite{achiam2023gpt}, Gemini series~\cite{google2025gemini25}, Qwen3~\cite{yang2025qwen3}, and Llama 4~\cite{meta2025llama4} have demonstrated remarkable multilingual capabilities. These models leverage massive pre-training datasets spanning dozens to hundreds of languages, such as the corpus used by Qwen3, encompassing 119 languages and dialects. However, research indicates persistent challenges in these systems, including the "English pivot" phenomenon~\cite{zhong2024englishcentricllmslanguagemultilingual} where models internally process non-English inputs through English-like representations, and consistent performance gaps between high-resource and low-resource languages. Our work with MMLU-ProX specifically addresses these challenges by providing a comprehensive evaluation framework that enables direct assessment of reasoning capabilities across linguistically diverse contexts.

\noindent \textbf{LLM Evaluation Benchmarks.}
Prior work on multilingual LLM evaluation has largely focused on breadth or translation fidelity, but often at the expense of reasoning depth or language nuance. Benchmarks like MMLU~\cite{hendrycks2021measuringmassivemultitasklanguage}, TurkishMMLU~\cite{yuksel-etal-2024-turkishmmlu} and KMMLU~\cite{son-etal-2025-kmmlu} evaluate expert reasoning tasks but are limited to a single language, while MGSM~\cite{shi2022languagemodelsmultilingualchainofthought} and XCOPA~\cite{ponti2020xcopa} prioritize multilingual coverage through translated or templated questions yet restrict evaluation to narrow reasoning formats such as math problems or causal inferences. Global-MMLU~\cite{singh2024global} extends MMLU to 42 languages with human-machine hybrid translations, but it suffers from inconsistent translation quality and remains limited in reasoning difficulty. MMLU-Pro~\cite{wang2024mmlu} extends the original MMLU benchmark by introducing highly complex reasoning questions and more distractor options better to evaluate LLMs' reasoning depth and robustness in English. Similarly, Humanity's Last Exam~\cite{phan2025humanity} is a rigorous benchmark of 3,000 expert-crafted questions across diverse subjects, designed to challenge advanced AI systems and assess their progress toward expert-level reasoning, but it still remains to be an English-centric benchmark.  While early comprehensive benchmarks such as XTREME~\cite{ruder2021xtreme} and XGLUE~\cite{liang2020xglue} significantly advanced the evaluation of cross-lingual transfer, they primarily focused on traditional tasks, often assessing generalization from English training data rather than deep LLM reasoning. This landscape underscores the need for benchmarks that not only cover diverse languages but also rigorously assess complex reasoning within appropriate cultural contexts, a gap that MMLU-ProX aims to address. A detailed comparison of the aforementioned dataset is shown in \autoref{tab:compare}. Among the selected benchmarks, MMLU-ProX fills an important gap by maintaining a balanced distribution of languages, subjects, and questions, with a focus on data parallelization and reasoning-focused features.
\section{Benchmark}

\subsection{Overview}
MMLU-ProX extends the challenging MMLU-Pro benchmark to encompass 29 typologically diverse languages: English (EN), Chinese (ZH), Japanese (JA), Korean (KO), French (FR), German (DE), Spanish (ES), Portuguese (PT), Arabic (AR), Thai (TH), Hindi (HI), Bengali (BN), Swahili (SW), Afrikaans (AF), Czech (CS), Hungarian (HU), Indonesian (ID), Italian (IT), Marathi (MR), Nepali (NE), Russian (RU), Serbian (SR), Telugu (TE), Ukrainian (UK), Urdu (UR), Vietnamese (VI), Wolof (WO), Yoruba (YO), and Zulu (ZU). MMLU-ProX benchmark maintains the high difficulty level and reasoning focus of MMLU-Pro while enabling rigorous evaluation of LLMs' cross-lingual reasoning capabilities. By carefully translating the same set of questions across all languages, MMLU-ProX facilitates direct comparison of model performance across linguistic boundaries while controlling for question difficulty.

\begin{figure}
    \centering
    \includegraphics[width=0.95\linewidth]{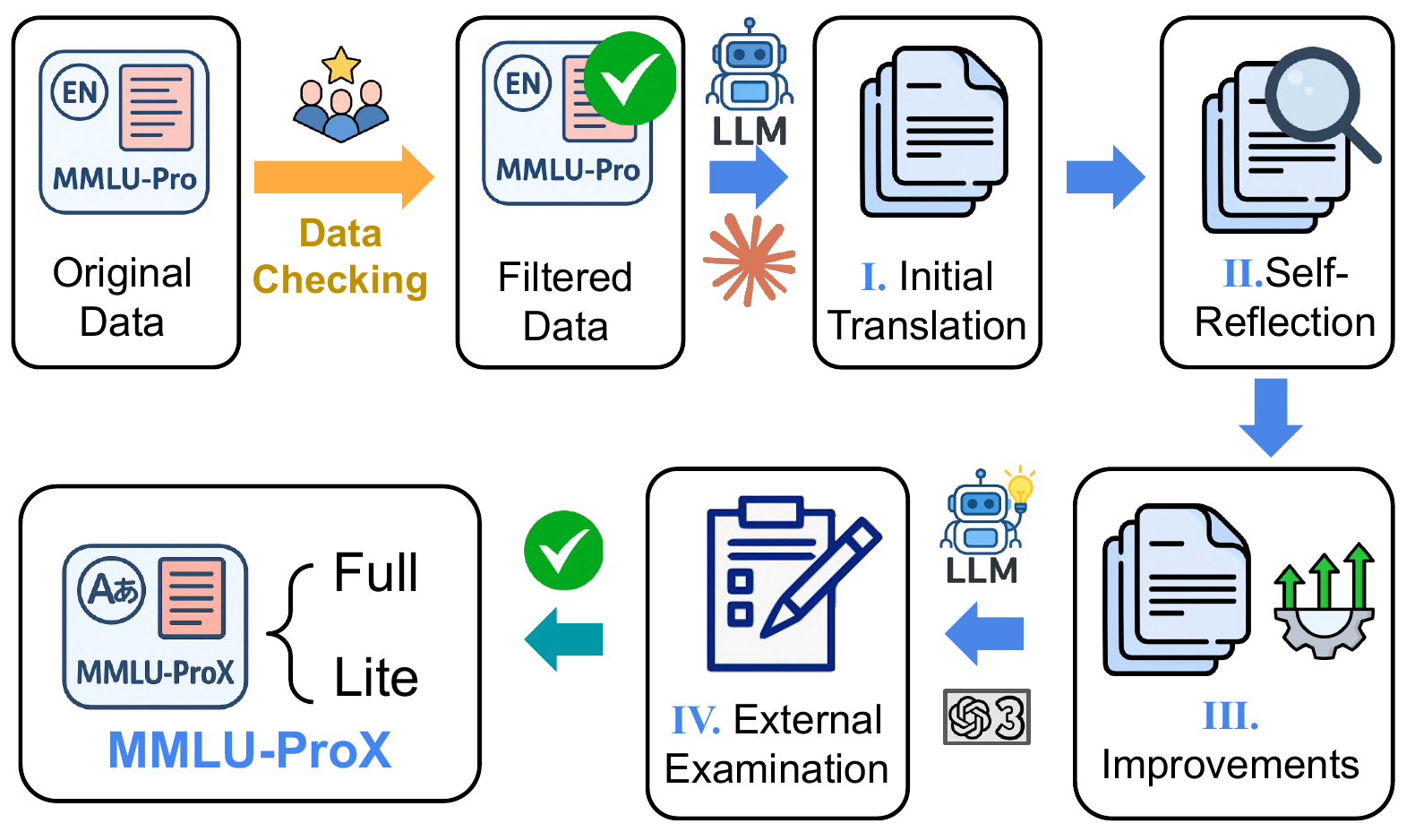}
    \caption{MMLU-ProX Data Pipeline: A rigorous four-stage process consisting of data curation, translation, external model verification, and expert review.}
    \label{fig:pipeline}
    \vspace{-5mm}
\end{figure}

To ensure the quality of MMLU-ProX, we implemented a multi-stage pipeline to generate the data shown in \autoref{fig:pipeline}. Initially, we hired a dedicated team to perform preliminary data curation, establishing a clean version suitable for subsequent translations. Following this, we deployed a translation agent to maintain translation quality standards. Finally, we employed a sampling methodology wherein professional translators evaluated selected samples. The results demonstrated that the generated dataset successfully passed assessment by professional human translators.

\subsection{Data Curation}
The data curation process comprises multiple stages. First, we identify and address duplicate or partially duplicate questions within MMLU-Pro, either eliminating or merging these instances to ensure fair evaluation without redundancy bias. Subsequently, we performed manual corrections of grammatical issues in the filtered dataset, addressing problems that include but are not limited to run-on words, incorrect hyphenation, and inconsistent symbol usage. Finally, we manually rectify inconsistencies within the questions, particularly focusing on misalignments between options and problem statements. For English data curation, we engage four interdisciplinary specialists, with a total labor investment of approximately 20 hours to correct run-on words, incorrect hyphenation, and other syntactic anomalies that could potentially confound the translation process. This curation step is critical to establish a clean baseline for our multilingual translations, as source-language errors can propagate and amplify through translation pipelines, particularly in technical and specialized domains that predominate in MMLU-Pro.

\subsection{Translation Pipeline}
Our data curation is followed by implementing a robust translation methodology. Based on recent machine translation evaluations~\cite{deutsch2025wmt24++, niklaus2025swiltra}, we select the SOTA Claude model, Claude Sonnet 3.7~\cite{anthropic2025claude} as our primary translation model. Although LLMs have shown impressive translation capabilities, we recognize the need to safeguard against potential translation errors. To address this, we develop a four-stage LLM-driven translation framework for producing MMLU-ProX:

\noindent \textbf{I. Initial Translation}: Claude Sonnet 3.7 performs the preliminary translations using carefully crafted prompts. These prompts emphasize maintaining accurate expression, consistent terminology across questions and options, and cultural appropriateness for target language users. The translation preserves all LaTeX notation, mathematical formulae, programming code (including variable names and comments), and currency symbols exactly as they appear in the source text. For units of measurement, we implement standard translations in target languages while maintaining precise numerical relationships and retaining all special formatting and emphasis from the original text.

\noindent \textbf{II. Self-Reflection}: In this stage, Claude Sonnet 3.7 performs a comprehensive review of its own translation's correspondence with the source text, generating feedback for improving the translation quality. The reflection process focuses on verifying proper noun translations and eliminating any superfluous explanations or additions. It also ensures the use of established technical terminology in the target language.

\noindent \textbf{III. Improvements}: Claude Sonnet 3.7 then conducts meticulous editing, incorporating feedback from the self-reflection stage. Additionally, we prompt the model to ensure the explanatory information is only included for concepts lacking direct equivalents in target languages, particularly in low-resource languages like Wolof and Yoruba. The LLM-driven process maintains strict preservation of original single quotation marks and removes any unnecessary explanations or source language terms.

\noindent \textbf{IV. External Examination}: To mitigate potential systematic errors from single-model biases, we employ two different LLMs for verification: OpenAI O3 for low-resource African languages (Swahili, Zulu, Yoruba, and Wolof), and GPT-4.1 for the rest. This automated verification process is designed to flag only significant discrepancies for manual review and human translation.

Appendix \S \ref{app:prompts} contains all translation prompts.
\subsection{Expert Verification}
\label{sec:human_verification}
To rigorously evaluate the benchmark quality following our translation framework implementation, we conduct comprehensive expert evaluations of the translation quality. Specifically, we randomly sample 20 items from each of the 14 disciplines and use these consistent items across all languages for evaluation. We select 15 languages and conduct expert verification, with each language evaluated by two professional translators who are native speakers of the target language and proficient in English. The total annotation effort exceeds 400 hours. Each item undergoes assessment by two high-caliber translators who rate three aspects—accuracy, fluency, and completeness—on a scale of 1–5. Detailed scoring criteria and full results can be found in Appendix \S\ref{app:human_ver}.

Subsequently, for any category where both translators assign scores below 3 on any metric, we conduct a complete retranslation of the entire discipline of that language to ensure final averaged scores for all categories strictly exceed 4 points, an indication of accurate translation. Throughout this process, only the law category in Yoruba requires such modification; all other categories across all languages maintain average scores of 4 or above.

\autoref{tab:human_verfication} presents representative translation scores, showing examples from three languages within each resource group (group criteria are in Appendix \S \ref{app:highlow}). The results demonstrate consistently high translation quality under expert evaluation, even for low-resource languages such as Wolof, Yoruba, and Nepali. This uniform performance across resource groups validates both the reliability of our translation pipeline and the overall quality of MMLU-ProX.

\begin{table}[ht]
\centering
\resizebox{0.45\textwidth}{!}{%
\begin{tabular}{lccc}
\toprule
\textbf{Language} & \textbf{Accuracy} & \textbf{Fluency} & \textbf{Completeness} \\
\midrule
\multicolumn{4}{l}{\textbf{High Resource}} \\
ZH & 4.70 & 4.84 & 4.92 \\
JA & 4.60 & 4.65 & 4.99 \\
FR & 4.68 & 4.64 & 4.94 \\
\midrule
\multicolumn{4}{l}{\textbf{Medium Resource}} \\
KO & 4.90 & 4.41 & 4.97 \\
PT & 4.79 & 4.77 & 4.99 \\
AF & 4.77 & 4.78 & 4.99 \\
\midrule
\multicolumn{4}{l}{\textbf{Low Resource}} \\
WO & 4.14 & 4.42 & 4.83 \\
YO & 4.06 & 4.56 & 4.95 \\
NE & 4.61 & 4.73 & 4.91 \\
\bottomrule
\end{tabular}
}
\caption{Scores (out of 5) assigned by human translators for Accuracy, Fluency, and Completeness, grouped by language resource level (We show representatives here).}
\label{tab:human_verfication}
\vspace{-3mm}
\end{table}

\subsection{Total Cost}
The development of MMLU-ProX requires substantial resource investment. Taking into account API costs for translation and testing, expert verification expenses, and computational resources, the total development cost approaches \$80,000 at market rates. This investment demonstrates our commitment to creating a high-quality, reliable benchmark for advancing multilingual LLM capabilities.

\begin{table*}[!t]
    \centering
    \resizebox{0.95\textwidth}{!}{
    \begin{tabular}{l|ccc|cc|c|c|c|c|c|c|cccc}
    \toprule
    \textbf{Language} &
    \textcolor{gpttextcolor}{\rotatebox{90}{\textbf{o4-mini}}} &
    \textcolor{gpttextcolor}{\rotatebox{90}{\textbf{GPT-4.1}}} &
    \textcolor{gpttextcolor}{\rotatebox{90}{\textbf{GPT-4o}}} &
    \textcolor{deepseektextcolor}{\rotatebox{90}{\textbf{DeepSeek-R1}}} &
    \textcolor{deepseektextcolor}{\rotatebox{90}{\textbf{DeepSeek-V3}}} &
    \textcolor{mistraltextcolor}{\rotatebox{90}{\textbf{Mistral-3.1-24B}}} &
    \textcolor{internlmtextcolor}{\rotatebox{90}{\textbf{InternLM3-8B}}} &
    \textcolor{ayatextcolor}{\rotatebox{90}{\textbf{Aya-32B}}} &
    \textcolor{phitextcolor}{\rotatebox{90}{\textbf{Phi4-14B}}} &
    \textcolor{llamatextcolor}{\rotatebox{90}{\textbf{Llama3.3-70B}}} &
    \textcolor{gemmatextcolor}{\rotatebox{90}{\textbf{Gemma3-27B}}} &
    \textcolor{qwentextcolor}{\rotatebox{90}{\textbf{Qwen3-32B}}} &
    \textcolor{qwentextcolor}{\rotatebox{90}{\textbf{Qwen3-32B-Think}}} &
    \textcolor{qwentextcolor}{\rotatebox{90}{\textbf{Qwen3-235B}}} &
    \textcolor{qwentextcolor}{\rotatebox{90}{\textbf{Qwen3-235B-Think}}} \\
    \midrule
    \rowcolor{overall}
    Overall (AVG) & 69.3 & 72.7 & 61.1 & \textbf{75.5} & 70.5 & 45.9 & 17.2 & 25.6 & 49.9 & 55.8 & 56.6 & 59.9 & 66.3 & 66.7 & 74.9 \\
    \midrule
    \rowcolor{westeuropean}
    English (EN) & 73.7 & 79.8 & 59.9 & 79.5 & 79.6 & 62.0 & 40.8 & 40.8 & 71.5 & 65.7 & 66.5 & 71.8 & 74.9 & 73.5 & \textbf{80.7} \\
    \rowcolor{westeuropean}
    French (FR) & 72.2 & 75.7 & 66.7 & \textbf{81.3} & 76.3 & 60.6 & 38.3 & 36.5 & 61.9 & 62.1 & 63.5 & 68.4 & 72.1 & 72.5 & 80.6 \\
    \rowcolor{westeuropean}
    German (DE) & 73.5 & 76.4 & 69.6 & 76.7 & 75.1 & 58.5 & 36.7 & 36.7 & 64.1 & 59.8 & 61.0 & 67.6 & 71.7 & 71.3 & \textbf{80.4} \\
    \rowcolor{westeuropean}
    Spanish (ES) & 74.7 & 77.8 & 68.6 & 80.2 & 76.9 & 59.4 & 36.3 & 35.4 & 59.6 & 61.5 & 63.0 & 68.7 & 72.8 & 73.2 & \textbf{80.7} \\
    \rowcolor{westeuropean}
    Portuguese (PT) & 74.1 & 77.0 & 67.9 & 78.0 & 75.7 & 60.0 & 36.1 & 30.1 & 61.7 & 61.4 & 63.2 & 69.1 & 72.7 & 73.1 & \textbf{80.5} \\
    \rowcolor{westeuropean}
    Italian (IT) & 73.9 & 78.2 & 62.9 & 79.9 & 75.9 & 59.6 & 34.7 & 34.5 & 60.2 & 67.0 & 64.4 & 69.4 & 73.5 & 73.7 & \textbf{80.9} \\
    \midrule
    \rowcolor{indoaryan}
    Hindi (HI) & 71.8 & 74.5 & 62.3 & 77.5 & 71.6 & 40.8 & 5.2 & 27.7 & 47.8 & 55.4 & 58.4 & 61.5 & 70.4 & 67.6 & \textbf{78.7} \\
    \rowcolor{indoaryan}
    Bengali (BN) & 70.1 & 72.2 & 62.8 & 66.6 & 69.8 & 32.0 & 3.8 & 14.0 & 34.1 & 50.1 & 55.5 & 57.1 & 66.4 & 67.7 & \textbf{77.8} \\
    \rowcolor{indoaryan}
    Urdu (UR) & 72.0 & 68.3 & 59.6 & \textbf{76.2} & 70.3 & 43.3 & 2.5 & 20.9 & 41.8 & 56.3 & 56.7 & 62.4 & 70.8 & 68.7 & 76.1 \\
    \rowcolor{indoaryan}
    Telugu (TE) & 69.1 & 65.9 & 51.3 & 71.9 & 67.6 & 29.3 & 6.4 & 7.2 & 24.1 & 47.9 & 55.9 & 51.0 & 70.3 & 66.7 & \textbf{77.9} \\
    \rowcolor{indoaryan}
    Marathi (MR) & 70.7 & 72.2 & 68.1 & 70.4 & 69.8 & 30.7 & 2.0 & 13.5 & 43.2 & 56.4 & 56.1 & 58.9 & 70.7 & 67.7 & \textbf{78.5} \\
    \rowcolor{indoaryan}
    Nepali (NE) & 71.5 & 74.2 & 61.3 & \textbf{78.9} & 69.3 & 32.9 & 1.5 & 14.5 & 36.0 & 52.8 & 56.8 & 59.7 & 70.7 & 67.8 & 78.1 \\

    \midrule
    \rowcolor{eastasian}
    Chinese (ZH) & 72.6 & 75.5 & 64.6 & \textbf{78.0} & 73.9 & 56.5 & 24.2 & 37.4 & 62.3 & 58.4 & 60.4 & 67.0 & 68.7 & 70.5 & 77.4 \\
    \rowcolor{eastasian}
    Japanese (JA) & 71.5 & 75.6 & 45.8 & 76.9 & 72.9 & 54.4 & 20.6 & 29.9 & 56.5 & 57.0 & 59.3 & 62.6 & 70.2 & 68.8 & \textbf{77.1} \\
    \rowcolor{eastasian}
    Korean (KO) & 73.2 & 75.4 & 57.9 & 76.7 & 70.7 & 52.3 & 20.0 & 34.4 & 58.2 & 54.5 & 57.8 & 65.5 & 71.2 & 69.6 & \textbf{78.3} \\
    \rowcolor{eastasian}
    Vietnamese (VI) & 73.4 & \textbf{76.7} & 70.4 & 76.3 & 75.4 & 53.4 & 5.3 & 30.9 & 57.1 & 65.2 & 61.1 & 68.5 & 72.4 & 71.4 & 72.6 \\
    \rowcolor{eastasian}
    Thai (TH) & 72.0 & 75.1 & 66.7 & \textbf{78.7} & 71.2 & 35.4 & 5.5 & 14.9 & 51.7 & 56.0 & 56.7 & 56.1 & 70.4 & 68.8 & 77.1 \\
    \rowcolor{eastasian}
    Indonesian (ID) & 73.8 & 75.6 & 66.1 & \textbf{81.3} & 75.8 & 55.5 & 31.6 & 23.1 & 63.9 & 65.5 & 62.6 & 68.5 & 73.4 & 72.5 & 79.9 \\
    \midrule
    \rowcolor{african}
    Arabic (AR) & 72.5 & 74.1 & 68.3 & 76.2 & 72.4 & 49.8 & 9.1 & 36.6 & 56.8 & 51.0 & 58.7 & 64.9 & 70.4 & 70.1 & \textbf{78.7} \\
    \rowcolor{african}
    Afrikaans (AF) & 73.5 & 77.2 & 65.3 & \textbf{80.9} & 72.9 & 53.3 & 27.6 & 29.7 & 57.8 & 62.7 & 62.0 & 65.9 & 72.4 & 71.1 & 80.6 \\
     \rowcolor{african}
    Swahili (SW) & 66.9 & 71.9 & 58.6 & \textbf{75.0} & 63.4 & 31.4 & 2.2 & 9.0 & 35.2 & 49.0 & 52.8 & 46.4 & 56.7 & 56.3 & 70.8 \\
    \rowcolor{african}
    Wolof (WO) & 24.1 & 43.2 & 24.3 & \textbf{58.6} & 47.3 & 17.0 & 0.6 & 1.5 & 8.1 & 28.5 & 8.8 & 26.1 & 26.6 & 26.6 & 36.9 \\
    \rowcolor{african}
    Yoruba (YO) & 54.9 & 53.4 & 44.3 & \textbf{57.0} & 47.7 & 13.5 & 0.6 & 3.9 & 23.1 & 31.6 & 32.4 & 25.7 & 18.8 & 40.2 & 49.3 \\
    \rowcolor{african}
    Zulu (ZU) & 61.2 & 65.0 & 55.3 & \textbf{67.3} & 53.7 & 17.0 & 2.2 & 14.5 & 11.5 & 33.6 & 40.7 & 17.9 & 35.2 & 46.2 & 46.4 \\
    \midrule
        \rowcolor{other}
    Russian (RU) & 62.0 & 71.2 & 62.0 & 76.4 & 74.9 & 59.2 & 26.1 & 36.7 & 65.2 & 61.1 & 62.5 & 68.0 & 69.1 & 72.9 & \textbf{77.0} \\
     \rowcolor{other}
    Ukrainian (UK) & 73.3 & 76.4 & 56.8 & 76.8 & 74.2 & 56.0 & 27.5 & 35.9 & 61.3 & 59.9 & 61.7 & 68.0 & 73.5 & 72.5 & \textbf{78.8} \\
    \rowcolor{other}
    Serbian (SR) & 72.6 & 76.9 & 70.6 & \textbf{80.9} & 72.9 & 53.9 & 28.9 & 27.4 & 50.7 & 63.0 & 61.7 & 67.2 & 72.3 & 71.1 & 80.2 \\
    \rowcolor{other}
    Czech (CS) & 73.5 & 77.5 & 70.1 & 76.8 & 74.7 & 55.1 & 0.0 & 34.5 & 63.2 & 63.8 & 62.6 & 67.7 & 72.8 & 71.8 & \textbf{80.5} \\
    \rowcolor{other}
    Hungarian (HU) & 72.6 & 76.6 & 63.0 & 79.1 & 71.4 & 48.7 & 22.2 & 29.1 & 59.4 & 59.7 & 59.8 & 65.9 & 71.1 & 70.1 & \textbf{79.8} \\
    \bottomrule
    \end{tabular}
    }
    \caption{Model performance (\%) on MMLU-ProX across 29 languages. Languages are grouped by linguistic families with distinct colors. Best result per language is in \textbf{bold}. Full tables can be found in Appendix \S \ref{app:all_res}.}
    \label{tab:cot_results}
    \vspace{-5mm}
\end{table*}

\section{Experiments}

\subsection{Setups}
We evaluate a comprehensive set of 36 SOTA LLMs on MMLU-ProX across 29 linguistically diverse languages. The evaluation includes both open-weight and proprietary LLMs, representing various architectures, parameter scales, and training paradigms. The open-weight LLMs include Qwen~\cite{qwq32b}, Llama~\cite{grattafiori2024llama}, DeepSeek~\cite{guo2025deepseekr1}, Phi4~\cite{abdin2024phi4}, Gemma3~\cite{team2025gemma}, Mistral~\cite{mistral2025small}, Aya~\cite{aryabumi2024aya}, and InternLM~\cite{cai2024internlm2}, while the proprietary LLMs comprise o4-mini, GPT4.1 and GPT4o. Following MMLU-Pro, we primarily employ 5-shot CoT prompting for model evaluation. All experiments were conducted using vLLM for inference on an H100 cluster. Our rough estimation indicates that the unified evaluation consumed over 10,000 GPU hours.

\subsection{Overall Performance}
We present a comparison in \autoref{tab:cot_results}, showing the CoT performance across all 29 languages and the average results of selected models, specifically the largest or best-performing model from each family (15 out of 36). We roughly group the languages by geography (stated in Appendix \S \ref{app:family}): \colorbox{westeuropean}{Western Europe}, \colorbox{indoaryan}{South Asia}, \colorbox{eastasian}{East Asia \& Southeast Asia}, \colorbox{african}{Africa} and \colorbox{other}{Eastern Europe}. 

Our evaluation of these LLMs reveals significant disparities in multilingual capabilities. DeepSeek-R1 demonstrates superior overall performance with an average of 75.2\% across all languages, followed by GPT-4.1 (72.7\%) and DeepSeek-V3 (70.5\%). The performance generally correlates with model scale and architecture sophistication, with larger models typically outperforming their smaller counterparts. Among open-weight models, Qwen3-235B-Think shows exceptional capabilities, achieving SOTA results in several languages. However, there remains a substantial performance gap between high-resource and low-resource languages, with some models showing accuracy as low as 0.6\% on certain African languages while achieving over 75\% on Western European languages. The full evaluation for all 36 LLMs and 0-shot settings could be found in Appendix \S \ref{app:all_res}. We conduct a more detailed analysis in the following.

\subsection{Impact of Reasoning Mode in Multilingual Performance}
\begin{figure}[htbp]
    \centering
    \includegraphics[width=1.0\linewidth]{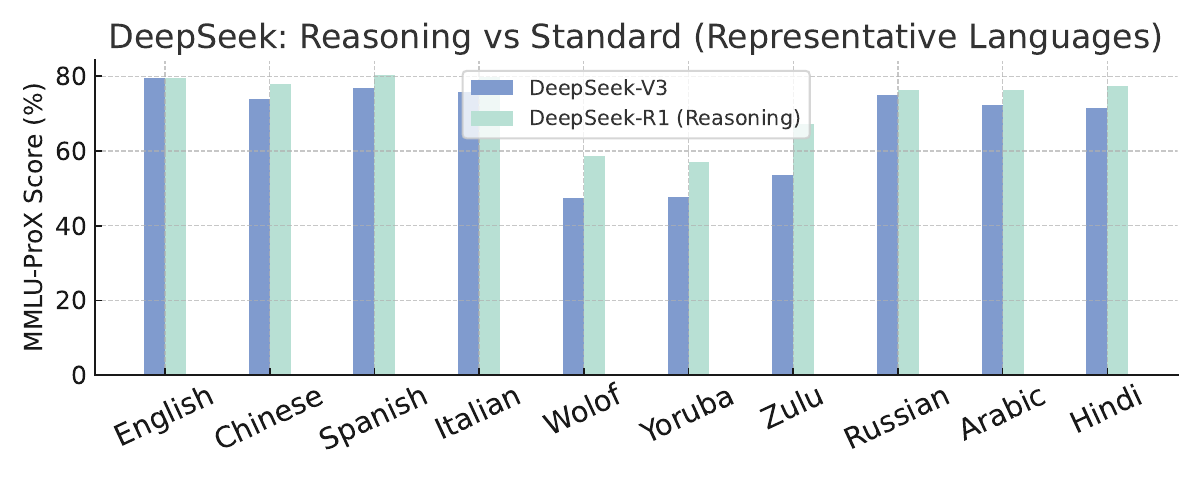}
    \vspace{1em}
    \includegraphics[width=1.0\linewidth]{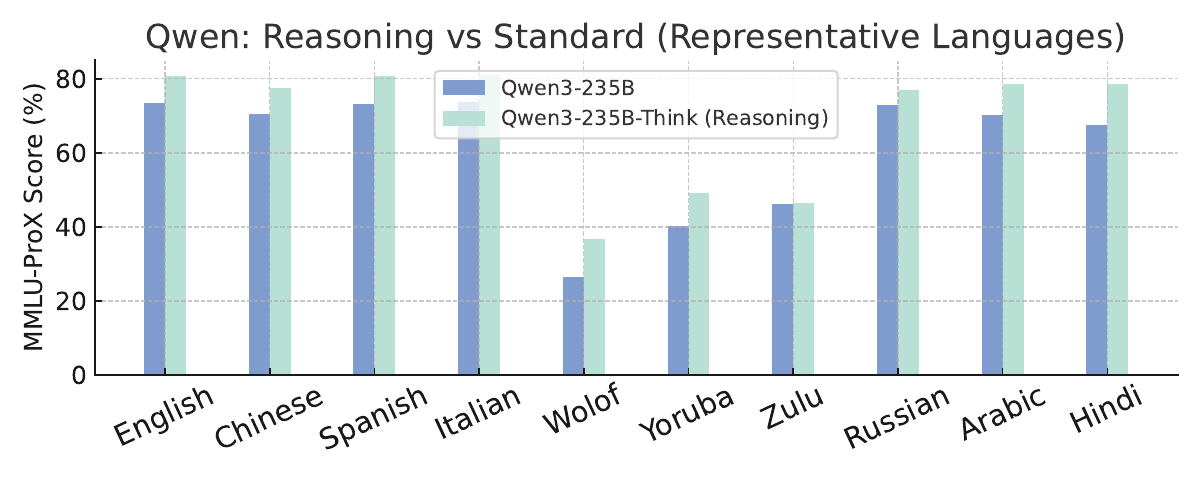}
    \caption{Comparison of reasoning-enhanced and standard models on representative languages. Top: DeepSeek-V3 vs DeepSeek-R1. Bottom: Qwen3-235B vs Qwen3-235B with thinking mode.}
    \label{fig:reasoning_bar_charts}
    \vspace{-3mm}
\end{figure}

We examine how reasoning-enhanced capabilities affect multilingual performance. Comparing reasoning-focused and standard LLMs reveals consistent performance improvements. DeepSeek-R1 outperforms DeepSeek-V3 by 4.7\% on average, with larger gains in low-resource languages (Wolof: +11.3\%, Yoruba: +9.3\%). Similarly, Qwen3-235B with thinking mode enabled achieves superior results compared to its base performance, reaching SOTA performance on Western European languages (English: 80.7\%, Spanish: 80.7\%, Italian: 80.9\%). These results suggest that reasoning-enhanced models better handle complex multilingual tasks, particularly in challenging linguistic contexts, indicating a promising direction for robust multilingual LLM development.

\begin{figure}[h]
    \centering
    \includegraphics[width=1.0\linewidth]{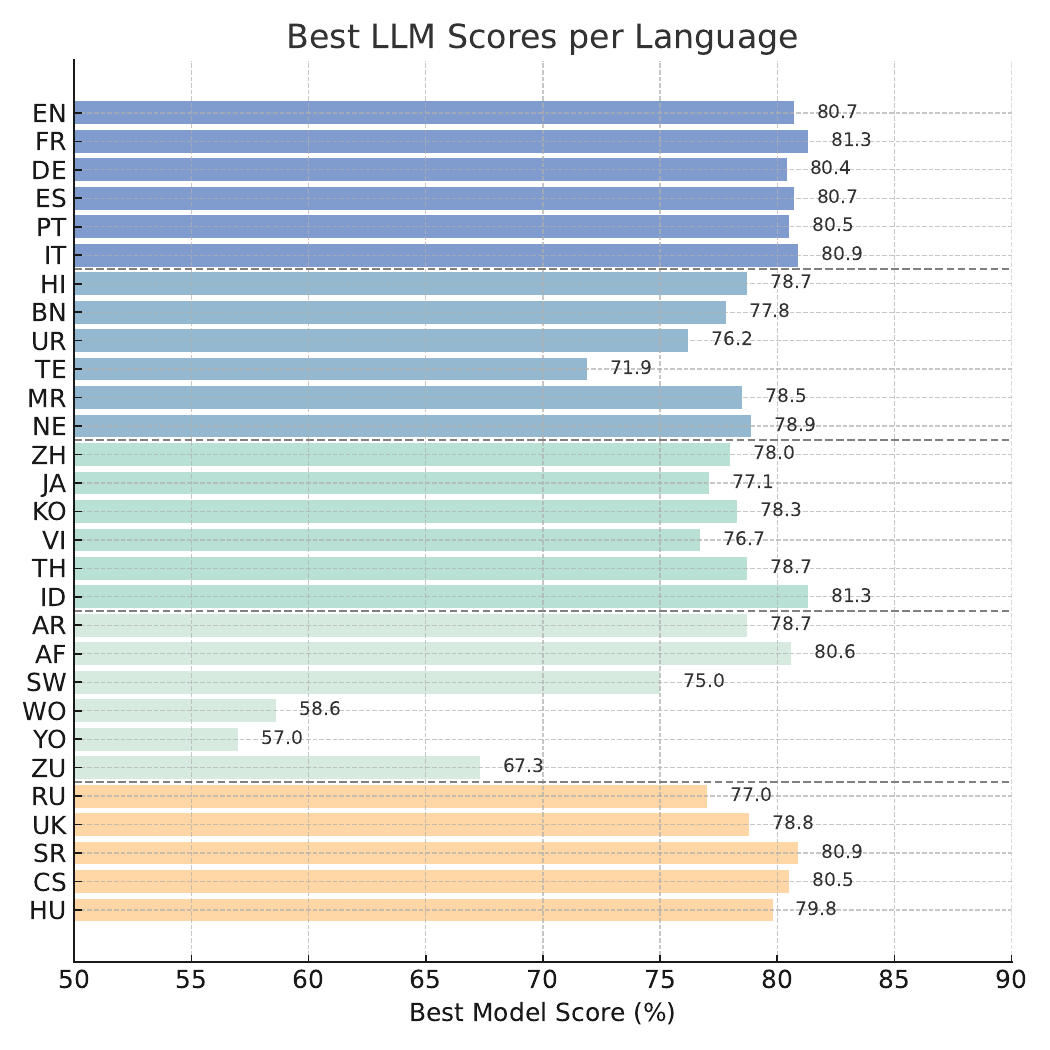}
    \caption{Best LLM scores on MMLU-ProX for each language, grouped by language family as in \autoref{tab:cot_results}.
    The figure highlights notable performance gaps between language families, especially the advantages for well-resourced Western and Eastern European languages compared to low-resource African and some South Asian languages.}
    \label{fig:language_family}
    \vspace{-5mm}
\end{figure}

\subsection{Performance across Language Groups}

The results in \autoref{tab:cot_results} reveal clear performance trends across linguistic groups and model families. Western European languages consistently achieve high accuracy across all models, with top-performing models (e.g., Qwen3-235B with thinking mode exceeds 77\% in every language in this group). South Asian languages show more variation: Hindi performs best within the group, while Telugu lags, highlighting challenges with Dravidian language modeling. DeepSeek-R1 and Qwen3-235B-Think stand out for their strong performance across several South Asian languages. East and Southeast Asian languages perform well overall, with Indonesian achieving 81.3\% and Japanese and Korean showing stable scores, despite linguistic divergence from Indo-European languages. In contrast, African languages demonstrate the lowest performance across the board. While Arabic performs competitively, other African languages such as Wolof, Yoruba, and Zulu exhibit wide performance gaps and significantly lower scores—Wolof ranging from just 0.6\% to 58.6\%—highlighting the persistent limitations of current models in low-resource settings. Notably, Eastern European languages also perform well, with models like DeepSeek and Qwen continuing to lead, suggesting effective adaptation to languages with linguistic similarity to Western European ones. More detailed analysis can be found in Appendix \S \ref{app:language_groups}.

\section{Analysis}
In this section, we present a detailed analysis and observations on model size and prompting strategies. We then compare the full and lite versions of MMLU-ProX. Finally, we analyze the translation pipeline using the Chinese and Japanese subsets.

\subsection{Model Size}

\begin{figure}[t]
    \centering
    \includegraphics[width=1.0\linewidth]{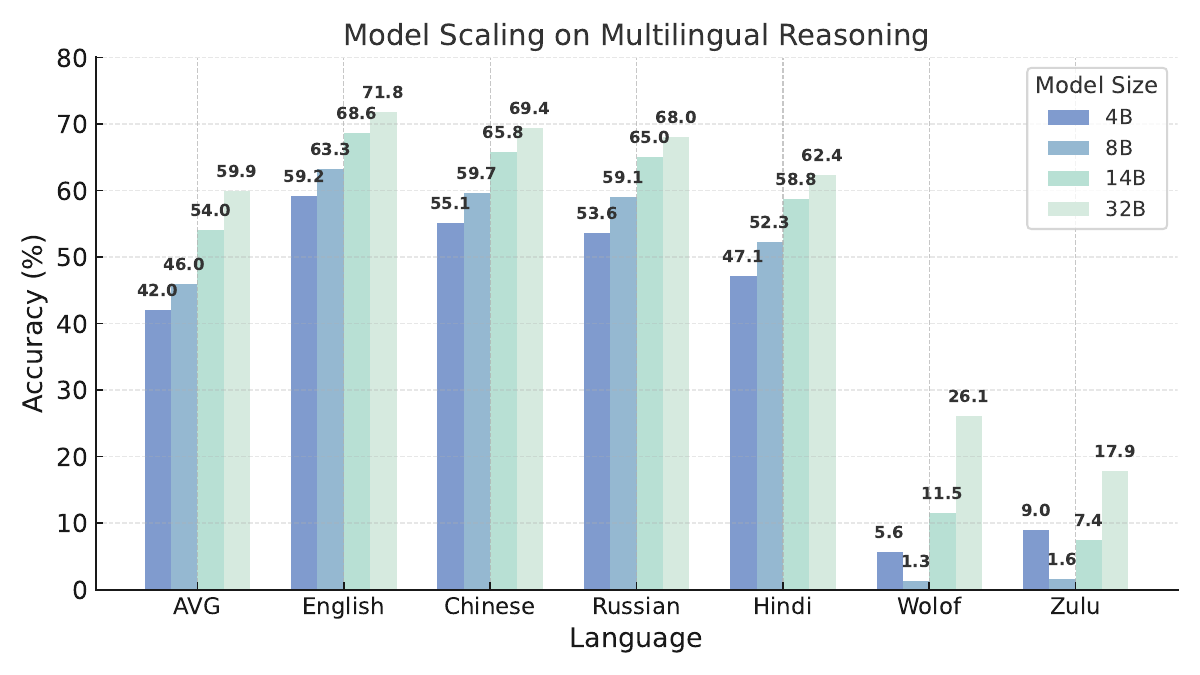}
    \caption{Performance scaling of Qwen3 dense models on selected languages.}
    \label{fig:size_scalling}
    \vspace{-3mm}
\end{figure}

In Figure~\ref{fig:size_scalling}, we analyze Qwen3 dense models at different scales (4B, 8B, 14B, 32B), revealing how model size affects multilingual reasoning. Performance improves consistently with scale, and the 32B model reaching 59.9\% accuracy—an absolute gain of 17.9\% over the 4B model on average. The largest improvement occurs from 8B to 14B (+8.0\%), while gains from 4B to 8B (+4.0\%) and 14B to 32B (+5.9\%) are more modest. High-resource languages like English show smaller differences (12.6\% from 59.2\% to 71.8\%), whereas low-resource languages benefit more: Wolof improves by 20.5\% and Russian by 14.4\%. In some African languages, such as Zulu, only the largest models show meaningful performance, suggesting a minimum model size may be required for effective multilingual capability.

These findings underscore how model scaling delivers asymmetric benefits across the linguistic spectrum, with low-resource languages typically requiring larger models to achieve even moderate performance. This differential scaling behavior highlights the importance of sufficient model capacity when developing multilingual systems intended to serve linguistically diverse user populations.

\subsection{Prompting Strategies}
\begin{figure*}[ht]
    \centering
    \includegraphics[width=1.0\linewidth]{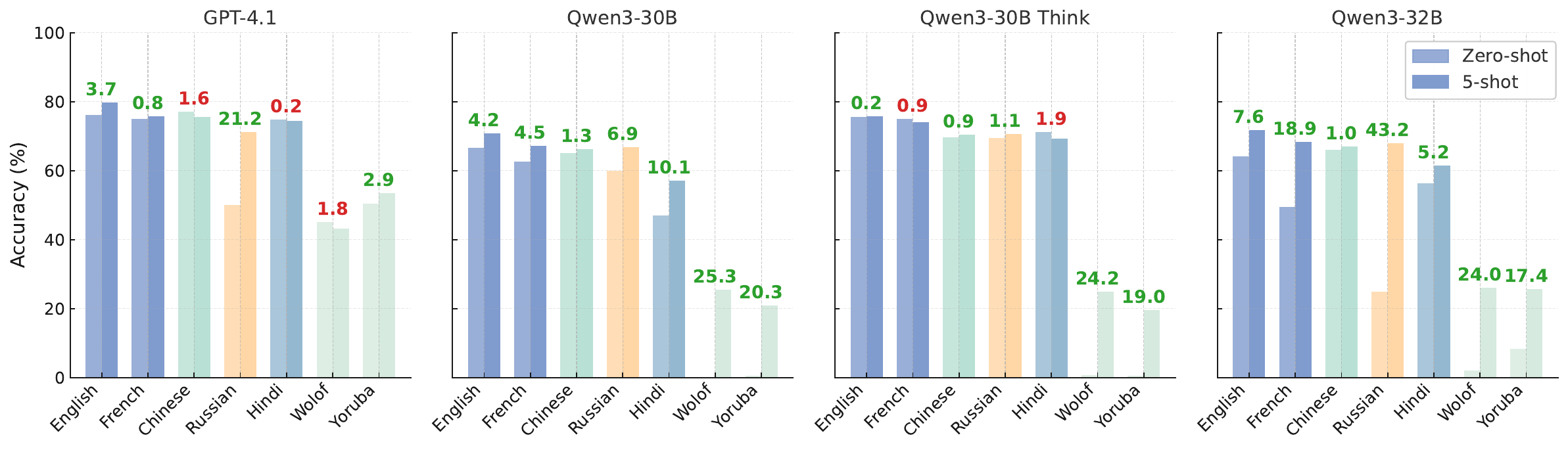}
    \caption{Performance comparison of zero-shot and 5-shot prompting across languages and models. The height of each bar represents accuracy (\%). Numbers above the bar pairs indicate the absolute difference in accuracy between 5-shot and zero-shot prompting for each language-model pair. Green numbers indicate improvement, while red numbers indicate a decrease.}
    \label{fig:prompting_strategies}
\end{figure*}

For LLM evaluation, prompting strategies play a crucial role. We selected representative LLMs, including GPT-4.1 and two Qwen variants, to comprehensively evaluate the effect of different prompting strategies on multilingual reasoning capabilities. We evaluate selected languages based on resource availability and linguistic families in \autoref{fig:prompting_strategies}. 

Our analysis reveals substantial performance differences between zero-shot and 5-shot prompting across languages and model families. While 5-shot prompting generally improves performance, the magnitude of improvement varies. High-resource languages like English show modest gains (e.g., +3.7\% for GPT-4.1), reflecting strong baseline reasoning abilities, whereas low-resource African languages benefit more significantly, indicating the added value of demonstrations in underrepresented languages. Reasoning-enhanced models such as Qwen3-30B in thinking mode show smaller changes between prompting styles, suggesting internalized reasoning capabilities. Additionally, language characteristics—such as morphological complexity—affect prompting effectiveness. These findings highlight the importance of tailoring prompting strategies to both model architecture and target language characteristics, particularly for multilingual applications targeting diverse linguistic environments.

\subsection{Full and Lite Versions}

\begin{figure}[h]
    \centering
    \includegraphics[width=1.0\linewidth]{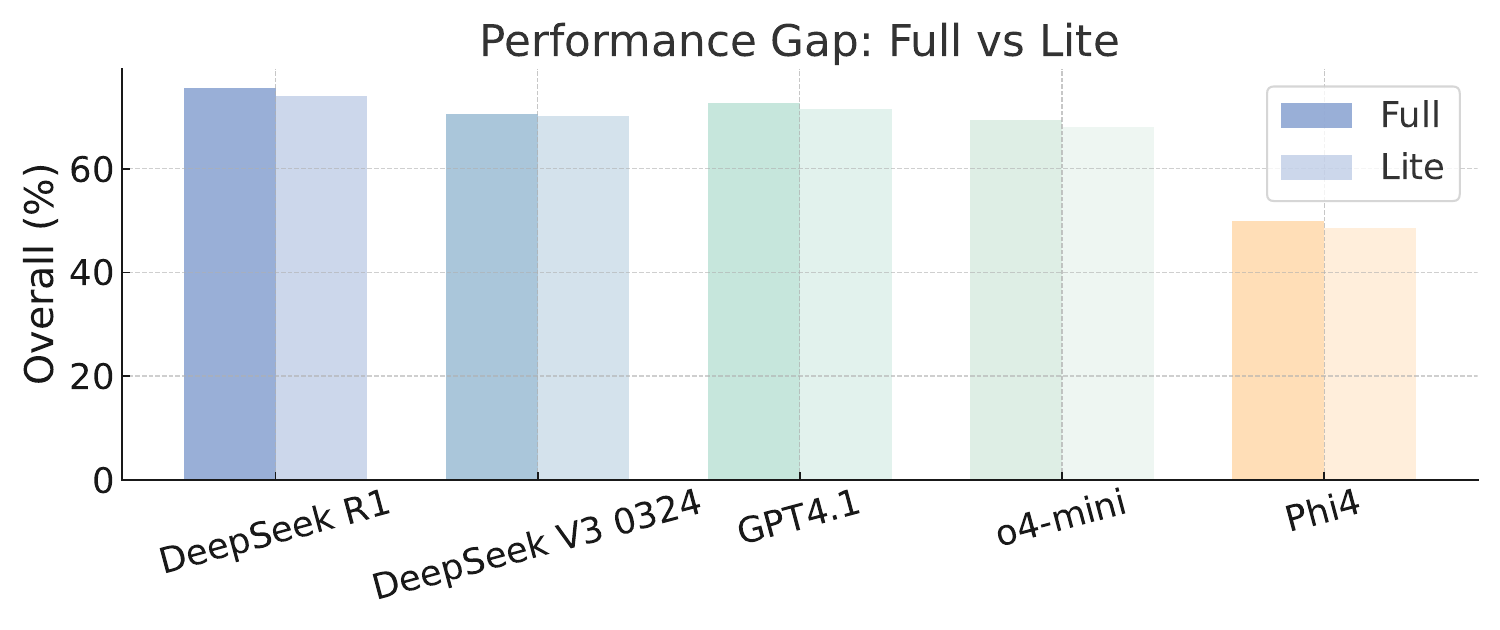}
    \caption{Comparison between MMLU-ProX full version (11,759 questions) and lite version (588 questions) across evaluated models, showing an average difference of only 1.14\% while maintaining consistent relative model rankings.}
    \label{fig:fullvslite}
\end{figure}

To address evaluation efficiency concerns in multilingual benchmarking, we also release a lite version by randomly sampling 20 items from each of the 14 disciplines—resulting in 658 items per language—for all 29 languages. We compare performance between the full version of MMLU-ProX (11,829 questions per language) and this lite version. Both versions include 70 validation questions used for prompt construction in few-shot evaluations, meaning actual assessments occur on 11,759 and 588 questions, respectively. As shown in \autoref{fig:fullvslite}, the performance gap between both versions is remarkably small, with an average difference of only 1.14\% across all evaluated models.

Across models and language families, the lite and full versions of MMLU-ProX yield highly consistent results. Top-performing models like DeepSeek-R1 and GPT-4.1 show minimal differences between lite and full evaluations (1.5\% and 1.1\%, respectively), with even smaller gaps observed in models like DeepSeek-V3 (0.4\%) and o4-mini (1.3\%). This pattern holds across resource levels, from high-resource languages like English and French to low-resource ones like Wolof, which shows differences under 1\%. Moreover, the lite version preserves the relative ranking of models almost perfectly, with DeepSeek-R1, GPT-4.1, and DeepSeek-V3 consistently outperforming others, while Phi4-14b remains the weakest. This consistent ordering confirms that the lite version effectively captures the same performance patterns as the full benchmark.

\section{Conclusion}
We introduce MMLU-ProX, a multilingual benchmark spanning 29 diverse languages for evaluating cross-lingual capabilities of LLMs. Our semi-automatic translation approach combines LLMs with expert verification to ensure quality across languages. Additionally, we conduct a comprehensive evaluation of 36 SOTA LLMs and reveal significant performance disparities in the multilingual setting. Our work aims to promote the equitable accessibility of LLMs in the global context.

\section*{Acknowledgments}
This project is supported by JST ACT-X (Grant JPMJAX24CU) and JSPS KAKENHI (Grant 24K20832). This work used supercomputers provided by the Research Institute for Information Technology, Kyushu University, through the HPCI System Research Project (Project ID: hp250092). This work is also supported by the NVIDIA Academic Grant Program and Google Cloud (Gemma 3 Academic Program).

\section*{Limitations}
In this work, we present MMLU-ProX, which covers 29 languages. One limitation lies in the coverage of languages due to budget constraints. While our current benchmark encompasses a diverse set of languages, expanding to include additional languages, particularly extremely low-resource ones, remains a future goal. We recognize that the existing pipeline can be extended to support such languages, and we leave this as future work.

Another limitation pertains to the expert verification of translation quality. While we engage experts to evaluate translation quality for selected languages, comprehensive expert verification across all languages and subject areas was not feasible due to resource constraints. In cases where expert evaluation was conducted, translations were assessed based on accuracy, fluency, and completeness using a 5-point Likert scale. Preliminary results indicate high overall quality, with mean scores above 4 across these dimensions. However, we acknowledge that automated translation processes may still introduce subtle errors or potential risks on the translation quality, particularly in complex or domain-specific content. 

Furthermore, the current benchmark focuses solely on textual inputs and does not account for multimodal contexts, which are increasingly relevant in real-world applications. Incorporating multimodal evaluation remains an area for future exploration.

\bibliography{custom,anthology_0, anthology_1}

\appendix

\onecolumn

\section{Language Categorization by Resource Availability}
\label{app:highlow}

Following the taxonomy proposed by~\cite{joshi-etal-2020-state}, and referring to the resource list\footnote{\url{https://microsoft.github.io/linguisticdiversity/assets/lang2tax.txt}}, we rank the languages from high- to low-resource as follows:\\Chinese, Japanese, French, German, Spanish, Arabic, Korean, Portuguese, Hindi, Serbian, Hungarian, Vietnamese, Czech, Italian, Russian, Thai, Bengali, Indonesian, Ukrainian, Urdu, Afrikaans, Zulu, Swahili, Wolof, Yoruba, Telugu, Marathi, Nepali.

\section{Language Categorization by Geography (\lq\lq by language family'')}
\label{app:family}

We primarily categorize the languages based on geography\footnote{\url{https://www.cia.gov/the-world-factbook/field/languages/}}. Inside each category, we rank the languages by resource abailabiliy. Below is the complete list of categories:

\begin{itemize}
    \item \textbf{Western Europe}: English, French, German, Spanish, Portuguese, Italian
    \item \textbf{South Asia}: Hindi, Bengali, Urdu, Telugu, Marathi, Nepali

    \item \textbf{East Asia \& Southeast Asia}: Chinese, Japanese, Korean, Vietnamese, Thai, Indonesian
        \item \textbf{Africa}: Arabic, Afrikaans, Swahili, Wolof, Yoruba, Zulu
    \item \textbf{Eastern Europe}: Russian, Ukrainian, Serbian, Czech, Hungarian

\end{itemize}

\section{Performance Patterns across Language Groups}
\label{app:language_groups}
The results in~\autoref{tab:cot_results} reveal distinct patterns across linguistic families, highlighting both achievements and persistent challenges in multilingual capabilities. 

As the Western European languages demonstrate consistently strong performance across all models, with scores typically ranging between 70-80\% for top-performing models. In this group, Qwen3-235B with thinking achieves remarkable results, reaching 80.9\% for Italian, 80.7\% for both English and Spanish, and maintaining above 77\% performance across all languages in this family.

South Asian languages exhibit a more nuanced performance pattern, with significant variations both across models and within the language family. Hindi consistently leads this group with scores ranging from 58.4\% to 78.7\%, while related languages like Bengali and Marathi show slightly lower but stable performance patterns. Telugu, representing the Dravidian family, generally shows lower performance across models, highlighting potential challenges in handling its distinct linguistic features. DeepSeek-R1 and Qwen3-235B-Think demonstrate particularly strong capabilities in this group, consistently achieving scores above 75\% for several languages.

East Asian \& Southeast Asian languages present an interesting case of high performance with model-specific variations. Chinese shows notable fluctuations across models (53.4-75.5\%), while Japanese and Korean demonstrate more consistent performance patterns. Southeast Asian languages perform remarkably well, with Indonesian achieving 81.3\% with DeepSeek-R1. This success suggests effective handling of these diverse linguistic structures by modern LLMs, despite the significant typological differences from Western languages.

African languages reveal the most pronounced performance disparities, underscoring critical challenges in multilingual AI development. While Arabic achieves competitive scores (up to 78.7\% with Qwen3-235B-Think), other African languages show substantially lower performance. Wolof presents the most challenging case, with scores ranging dramatically from 0.6\% to 58.6\%, highlighting severe resource limitations. Similar patterns emerge for Yoruba (3.9-57.0\%) and Zulu (11.5-67.3\%), though with slightly better performance than Wolof. These stark contrasts emphasize the ongoing need for improved model capabilities in low-resource languages.

Notably, Eastern European languages exhibit comparable performance. Similarly, models from the DeepSeek and Qwen families continue to perform strongly, with Qwen3-235B-Think achieving over 77\%. These strong performances suggest effective adaptation to these languages, likely due to their shared linguistic structures with Western European languages.

\section{Translation Pipeline Analysis}
\label{app:translation}
For our translation framework evaluation, we compared two competitors in English-to-Japanese translation: reasoning-based translation and human translators. Using the same samples as in Section~\ref{sec:human_verification}, we conducted translations using these two methods and employed professional translators to score them using our 5-point scale. The results are presented in \autoref{tab:en2ja}.
Our findings reveal that the reasoning-based method achieves translation quality only marginally inferior to our translation framework. However, compared to our agent-based method, reasoning-based translation consumes significantly more tokens, causing higher translation costs. As for native-speaking translators, their translation quality, particularly accuracy, proved inferior to LLM-based translation when handling content requiring multidisciplinary expertise. These results demonstrate the effectiveness of our comprehensive framework and further validate the quality of MMLU-ProX data.

\begin{table}[ht]
\centering
\resizebox{0.6\textwidth}{!}{%
\begin{tabular}{l|c|c|c}
\toprule
\rowcolor[gray]{0.95} \textbf{Method} & \textbf{Accuracy} & \textbf{Fluency} & \textbf{Completeness} \\
\midrule
\textbf{Agent-based Translation (Ours)} & \cellcolor[gray]{0.9}\textbf{4.60} & \cellcolor[gray]{0.9}\textbf{4.65} & \cellcolor[gray]{0.9}\textbf{4.99} \\
\hline
\textbf{Reasoning-based Translation} & 4.56 & 4.21 & \textbf{4.99} \\
\hline
\textbf{Native-Speaking Translator} & 4.24 & 4.14 & 4.56 \\
\bottomrule
\end{tabular}
}
\caption{Scores (out of 5) assigned by human translators for Accuracy, Fluency, and Completeness, grouped by language resource level.}
\label{tab:en2ja}
\end{table}

\section{Expert Verification Guidance}
\label{app:human_ver}
We ensured that all expert annotators were compensated at rates above the minimum hourly wage in their respective countries.
Evaluation Criteria for Expert Rating of Machine Translation Results:\\
\textbf{1. Accuracy (1-5):}

\begin{itemize}
    \item \textbf{5 (Highly Accurate):}
    \begin{itemize}
        \item All key terms and concepts are translated correctly with no errors.
        \item Every technical term corresponds precisely to the original text, with no mistranslations or incorrect word choices.
        \item The most appropriate and professional terminology in the target language is used.
        \item Expressions align with commonly used terminology in professional or technical contexts.
    \end{itemize}

    \item \textbf{4 (Accurate):}
    \begin{itemize}
        \item Most terms and concepts are translated correctly, with only a few minor errors that do not affect overall comprehension.
        \item Some terms may be slightly imprecise, but the translation remains generally accurate.
        \item Uses appropriate terminology in the target language in most cases.
        \item A few terms may be simplified but remain understandable within the intended domain.
    \end{itemize}

    \item \textbf{3 (Moderately Accurate):}
    \begin{itemize}
        \item Key terms and concepts are mostly correct but contain some errors that may cause partial misunderstandings.
        \item Some critical terms are inaccurately translated, requiring the reader to infer the intended meaning.
        \item Slight deviations in the use of target-language terminology.
        \item Occasionally uses uncommon or outdated terms.
    \end{itemize}

    \item \textbf{2 (Somewhat Inaccurate):}
    \begin{itemize}
        \item Many key terms and concepts are mistranslated, significantly affecting comprehension.
        \item Important concepts are incorrectly translated, leading to potential misunderstandings of the original text.
        \item Uses incorrect or inappropriate terminology in the target language.
        \item Terminology is inconsistent, reducing the text’s professionalism.
    \end{itemize}

    \item \textbf{1 (Inaccurate):}
    \begin{itemize}
        \item Frequent and severe mistranslations of key terms and concepts, failing to convey the original meaning.
        \item Most of the content does not match the original text.
        \item Lacks proper use of target-language terminology.
        \item Terminology is chaotic, possibly using irrelevant or incorrect vocabulary entirely.
    \end{itemize}
\end{itemize}

\textbf{2. Fluency (1–5):}

\begin{itemize}
    \item \textbf{5 (Highly Fluent):}
    \begin{itemize}
        \item The target-language expression is natural and smooth, making it effortless to read.
        \item The language style is refined and appropriate for professional or formal contexts.
        \item The sentence structure fully adheres to natural conventions in the target language, with no grammatical or lexical errors.
    \end{itemize}

    \item \textbf{4 (Fluent):}
    \begin{itemize}
        \item The target-language expression is generally natural, with only minor linguistic imperfections that do not affect comprehension.
        \item Some sentences may sound slightly stiff.
        \item Sentence structures mostly conform to target-language norms, with very few grammatical errors.
    \end{itemize}

    \item \textbf{3 (Moderately Fluent):}
    \begin{itemize}
        \item The target-language expression is somewhat unnatural, requiring the reader to adjust their understanding slightly.
        \item Some inappropriate word choices or rigid sentence structures are present.
        \item Sentence structures are mostly correct, but some grammatical errors exist.
    \end{itemize}

    \item \textbf{2 (Somewhat Unnatural):}
    \begin{itemize}
        \item The target-language expression lacks fluency, making it difficult to read smoothly.
        \item Sentence transitions are awkward, and logical connections are unclear.
        \item Many structural issues exist, with frequent grammatical errors.
    \end{itemize}

    \item \textbf{1 (Not Fluent):}
    \begin{itemize}
        \item The target-language expression is highly unnatural or difficult to understand.
        \item Literal translation is evident, lacking natural phrasing in the target language.
        \item The sentence structure is disorganized, with severe grammatical mistakes, making the text unreadable.
    \end{itemize}
\end{itemize}

\textbf{3. Completeness (1–5):}

\begin{itemize}
    \item \textbf{5 (Fully Complete):}
    \begin{itemize}
        \item The full meaning of the original text is retained with no omissions or additions.
        \item All details, data, and annotations are accurately conveyed.
        \item The translation maintains the same length and depth as the original text.
    \end{itemize}

    \item \textbf{4 (Complete):}
    \begin{itemize}
        \item The primary meaning of the original text is retained, with only a few minor details omitted or slightly unclear.
        \item Some less critical information may be left out.
        \item The translation generally corresponds to the original content.
    \end{itemize}

    \item \textbf{3 (Moderately Complete):}
    \begin{itemize}
        \item Most of the original meaning is conveyed, but some information is missing or added.
        \item Important details may be overlooked.
        \item The translation differs from the original in certain aspects, requiring readers to infer some content.
    \end{itemize}

    \item \textbf{2 (Somewhat Incomplete):}
    \begin{itemize}
        \item The core information from the original text is not fully conveyed, with noticeable omissions or unnecessary additions.
        \item Potential inclusion of unrelated information.
        \item The translation does not fully correspond to the original, affecting comprehension.
    \end{itemize}

    \item \textbf{1 (Incomplete):}
    \begin{itemize}
        \item Significant omissions or added incorrect information prevent an accurate reflection of the original text.
        \item Important sections or sentences are missing.
        \item The translation deviates heavily from the original, making it difficult to understand the intended meaning.
    \end{itemize}
\end{itemize}

\textbf{Scoring Examples:}

\begin{itemize}
    \item \textbf{Accuracy Example:} \\
    If ``bachelor's degree'' is mistranslated as ``single man's degree,'' points should be deducted in the accuracy category.

    \item \textbf{Fluency Example:} \\
    If the sentence structure follows target-language norms but the word choice is slightly unnatural, a score of 4 may be appropriate.

    \item \textbf{Completeness Example:} \\
    If the translated text omits the methodology section from the original, it should receive a lower score in completeness.
\end{itemize}

We perform expert verification in 15 selected languages, the full results are shown in \autoref{tab:human_verfication_full}. 

\begin{table}[h]
\centering
\resizebox{0.45\textwidth}{!}{%
\begin{tabular}{lccc}
\toprule
\textbf{Language} & \textbf{Accuracy} & \textbf{Fluency} & \textbf{Completeness} \\
\midrule
\multicolumn{4}{l}{\textbf{High Resource}} \\
ZH & 4.70 & 4.84 & 4.92 \\
JA & 4.60 & 4.65 & 4.99 \\
FR & 4.68 & 4.64 & 4.94 \\
DE & 4.52 & 4.48 & 4.64 \\
ES & 4.59 & 4.58 & 4.84 \\
\midrule
\multicolumn{4}{l}{\textbf{Medium Resource}} \\
KO & 4.90 & 4.41 & 4.97 \\
PT & 4.79 & 4.77 & 4.99 \\
AF & 4.77 & 4.78 & 4.99 \\
\midrule
\multicolumn{4}{l}{\textbf{Low Resource}} \\
ZU & 4.20 & 4.62 & 4.97 \\
SW & 4.36 & 4.70 & 4.86 \\
WO & 4.14 & 4.42 & 4.83 \\
YO & 4.06 & 4.56 & 4.95 \\
TE & 4.60 & 4.74 & 4.97 \\
MR & 4.51 & 4.72 & 4.74 \\
NE & 4.61 & 4.73 & 4.91 \\
\bottomrule
\end{tabular}
}
\caption{Expert verification scores on 15 languages for Accuracy, Fluency, and Completeness, grouped by language resource level.}
\label{tab:human_verfication_full}
\end{table}

\newpage
\section{Translation Prompts}
\label{app:prompts}

\begin{tcolorbox}[remarkbox, title=initial\_translation]
\small
\textbf{System Message:} You are a professional translator specializing in accurate translation of technical and academic content from \{source\_lang\} to \{target\_lang\}.\\
\\
Your task is to translate assessment questions in the \{category\} field while:\\
1. Preserving technical accuracy and terminology\\
2. Ensuring cultural appropriateness for \{target\_lang\} speakers\\
3. Keeping terminology consistent throughout questions and options\\
4. Preserving all LaTeX notation, mathematical formulas, and programming code exactly as they appear (do not translate content inside LaTeX delimiters or code blocks, including variable names, function names, and comments)\\
5. Preserving all currency symbols (\$) exactly as they appear in the original text, without converting to local currency units\\
6. For units of measurement: Use the conventional translations in the target language while preserving the exact numerical values and relationships\\
7. Preserving any special formatting or emphasis in the original text\\
\\
Please translate the following \{category\} assessment question and options:\\
<SOURCE\_TEXT>\\
\{source\_text\}\\
</SOURCE\_TEXT>\\
\\
\textbf{Output:}\\
Only provide the \{target\_lang\} translation for the above text. Do not include any explanations or text apart from the translation.\\
Different options are separated by newline characters(\verb|\n|).\\
The number of options in the output must match the input exactly. Do not skip or combine any options.\\
Return the translation in the following JSON format, with keys "question" and "options", where the value of "options" is a dictionary with keys option1, option2, option3, etc. All JSON keys must remain in English exactly as shown and only translate the content inside square brackets:\\
\\
<TRANSLATION>\\
\{\{\\
\quad "question": "[translation of question]",\\
\quad "options": \{\{\\
\quad\quad "option1": "\char91 translation of option1 \char93",\\
\quad\quad "option2": "\char91 translation of option2 \char93",\\
\quad\quad "option3": "\char91 translation of option3 \char93",\\
\quad\quad ...\\
\quad \}\}\\
\}\}\\
</TRANSLATION>
\end{tcolorbox}

\subsection{Self-Reflection Prompt}

\begin{tcolorbox}[remarkbox, title=self\_reflection]
\small
\textbf{System Message:} You are a \{category\} translation expert, specializing in translation from \{source\_lang\} to \{target\_lang\}.\\
\\
\textbf{Task Description:}\\
Carefully review the source text and its translation from \{source\_lang\} to \{target\_lang\}, and then provide constructive suggestions in English.\\
\\
\textbf{Requirements:}\\
1. Do not add, remove, or explain any information.\\
2. Make sure retain the original format for specialized information, e.g., anonymous information.\\
3. Identify any instances where proper nouns remain untranslated or where the translation contains unnecessary explanations, parenthetical original terms, or additions from \{source\_lang\}.\\
4. Examine whether any technical terms, subject-specific concepts, or other specialized vocabulary have been left in \{source\_lang\} instead of using their established \{target\_lang\} equivalents.\\
5. Verify that currency symbols, mathematical operators, and measurement units remain exactly as they appear in \{source\_lang\} text. These symbols should not be converted to their written form in \{target\_lang\}.\\
6. Check that no additional symbols or written representations have been added to options where they did not exist in \{source\_lang\} text.\\
\\
\textbf{Input:}\\
<SOURCE\_TEXT>\\
\{source\_text\}\\
</SOURCE\_TEXT>\\
\\
<INITIAL\_TRANSLATION>\\
\{initial\_trans\}\\
</INITIAL\_TRANSLATION>\\
\\
\textbf{Output:}\\
<SUGGESTIONS>\\
\char91 Your suggestions here \char93\\
</SUGGESTIONS>
\end{tcolorbox}

\subsection{Translation Improvement Prompt}

\begin{tcolorbox}[remarkbox, title=improve\_translation]
\small
\textbf{System Message:} You are a \{category\} translation expert, specializing in translation from \{source\_lang\} to \{target\_lang\}.\\
\\
\textbf{Task Description:}\\
Carefully review and edit the \{category\} translation from \{source\_lang\} to \{target\_lang\}, incorporating the expert feedback.\\
\\
\textbf{Requirements:}\\
1. Do not explain any information.\\
2. Strictly keep the single quotes in the original text and do not add new single and double quotes.\\
3. Remove unnecessary explanations or original terms from \{source\_lang\} if present in the translation.\\
\\
\textbf{Input:}\\
<SOURCE\_TEXT>\\
\{source\_text\}\\
</SOURCE\_TEXT>\\
\\
<INITIAL\_TRANSLATION>\\
\{initial\_trans\}\\
</INITIAL\_TRANSLATION>\\
\\
<EXPERT\_SUGGESTIONS>\\
\{reflection\}\\
</EXPERT\_SUGGESTIONS>\\
\\
\textbf{Output:}\\
Only provide the improved translation. Do not include any explanations or text apart from the translation.\\
Different options are separated by newline characters(\verb|\n|).\\
The number of options in the output must match the input exactly. Do not skip or combine any options.\\
Return the translation in the following JSON format, with keys "question" and "options", where the value of "options" is a dictionary with keys option1, option2, option3, etc. All JSON keys must remain in English exactly as shown and only translate the content inside square brackets:\\
\\
<IMPROVED\_TRANSLATION>\\
\{\{\\
    "question": "\char91 improved translation of question \char93",\\
    "options": \{\{\\
        "option1": "\char91 improved translation of option1 \char93",\\
        "option2": "\char91 improved translation of option2 \char93",\\
        "option3": "\char91 improved translation of option3 \char93",\\
        ...\\
    \}\}\\
\}\}\\
</IMPROVED\_TRANSLATION>\\
\end{tcolorbox}

\section{Detailed Evaluation Results}
\label{app:all_res}
We put more results in \autoref{app:full_res1}, \autoref{app:full_res2}, \autoref{app:full_res3}, and \autoref{app:full_res4}.

\onecolumn
\begin{sidewaystable}
    \small %
    \setlength{\tabcolsep}{3pt} 
    \centering

    \begin{longtable}{l|c|cccccc|cccccc|cccccc|cccccc|ccccc}
    \toprule
    Model & \textbf{Avg} & \textbf{EN} & \textbf{FR} & \textbf{DE} & \textbf{ES} & \textbf{PT} & \textbf{IT} & \textbf{HI} & \textbf{BN} & \textbf{UR} & \textbf{TE} & \textbf{MR} & \textbf{NE} & \textbf{ZH} & \textbf{JA} & \textbf{KO} & \textbf{VI} & \textbf{TH} & \textbf{ID} & \textbf{AR} & \textbf{AF} & \textbf{SW} & \textbf{WO} & \textbf{YO} & \textbf{ZU} & \textbf{RU} & \textbf{UK} & \textbf{SR} & \textbf{CS} & \textbf{HU} \\
    \midrule
    
    \rowcolor{gptcolor}
    o4-mini & 69.3 & 73.7 & 72.2 & 73.5 & 74.7 & 74.1 & 73.9 & 71.8 & 70.1 & 72.0 & 69.1 & 70.7 & 71.5 & 72.6 & 71.5 & 73.2 & 73.4 & 72.0 & 73.8 & 72.5 & 73.5 & 66.9 & 24.1 & 54.9 & 61.2 & 62.0 & 73.3 & 72.6 & 73.5 & 72.6 \\
    \rowcolor{gptcolor}
    GPT-4o & 61.1 & 59.9 & 66.7 & 69.6 & 68.6 & 67.9 & 62.9 & 62.3 & 62.8 & 59.6 & 51.3 & 68.1 & 61.3 & 64.6 & 45.8 & 57.9 & 70.4 & 66.7 & 66.1 & 68.3 & 65.3 & 58.6 & 24.3 & 44.3 & 55.3 & 62.0 & 56.8 & 70.6 & 70.1 & 63.0 \\
    \rowcolor{gptcolor}
    GPT-4.1 & 72.7 & 79.8 & 75.7 & 76.4 & 77.8 & 77.0 & 78.2 & 74.5 & 72.2 & 68.3 & 65.9 & 72.2 & 74.2 & 75.5 & 75.6 & 75.4 & \textbf{76.7} & 75.1 & 75.6 & 74.1 & 77.2 & 71.9 & 43.2 & 53.4 & 65.0 & 71.2 & 76.4 & 76.9 & 77.5 & 76.6 \\
    \rowcolor{deepseekcolor}
    DeepSeek-R1-7B & 17.2 & 28.9 & 31.8 & 31.9 & 38.7 & 37.3 & 39.6 & 4.7 & 0.3 & 3.3 & 2.5 & 1.9 & 5.6 & 33.5 & 2.6 & 3.3 & 33.1 & 10.8 & 40.4 & 28.8 & 23.1 & 0.9 & 0.3 & 0.0 & 0.3 & 18.4 & 24.2 & 26.3 & 25.3 & 0.7 \\
    \rowcolor{deepseekcolor}
    DeepSeek-R1-8B & 18.4 & 36.2 & 28.5 & 33.6 & 33.9 & 25.4 & 31.4 & 21.4 & 0.0 & 4.7 & 0.0 & 21.5 & 8.2 & 24.5 & 21.3 & 10.6 & 26.9 & 21.1 & 34.2 & 23.2 & 20.1 & 3.2 & 0.5 & 0.4 & 0.6 & 26.9 & 26.1 & 21.9 & 21.9 & 6.4 \\
    \rowcolor{deepseekcolor}
    DeepSeek-R1-14B & 33.7 & 59.2 & 53.6 & 48.1 & 58.8 & 50.9 & 55.0 & 21.0 & 28.7 & 4.3 & 4.4 & 4.1 & 7.1 & 33.1 & 45.8 & 21.0 & 55.8 & 57.5 & 53.3 & 53.2 & 22.0 & 16.9 & 22.0 & 0.2 & 0.1 & 20.9 & 59.7 & 33.1 & 42.5 & 11.2 \\
    \rowcolor{deepseekcolor}
    DeepSeek-R1-32B & 45.2 & 71.1 & 61.9 & 58.3 & 67.0 & 58.3 & 61.5 & 43.7 & 42.5 & 39.2 & 7.7 & 21.4 & 34.9 & 55.2 & 57.9 & 43.7 & 62.9 & 58.8 & 62.7 & 54.4 & 39.2 & 9.3 & 40.9 & 0.1 & 0.5 & 60.3 & 61.6 & 64.9 & 57.8 & 38.6 \\
    \rowcolor{deepseekcolor}
    DeepSeek-R1-70B & 58.7 & 70.3 & 73.7 & 72.5 & 73.7 & 70.6 & 73.6 & 62.0 & 49.3 & 56.9 & 41.2 & 48.4 & 44.0 & 53.4 & 59.8 & 47.4 & 69.3 & 60.0 & 71.5 & 62.6 & 73.1 & 62.0 & 12.2 & 26.3 & 46.7 & 52.7 & 57.5 & 67.1 & 71.7 & 71.5 \\
    \rowcolor{deepseekcolor}
    DeepSeek-R1 & \textbf{75.5} & 79.5 & \textbf{81.3} & 76.7 & 80.2 & 78.0 & 79.9 & 77.5 & 66.6 & \textbf{76.2} & 71.9 & 70.4 & \textbf{78.9} & \textbf{78.0} & 76.9 & 76.7 & 76.3 & \textbf{78.7} & \textbf{81.3} & 76.2 & \textbf{80.9} & \textbf{75.0} & \textbf{58.6} & \textbf{57.0} & \textbf{67.3} & 76.4 & 76.8 & \textbf{80.9} & 76.8 & 79.1 \\
    \rowcolor{deepseekcolor}
    DeepSeek-V3 & 70.5 & 79.6 & 76.3 & 75.1 & 76.9 & 75.7 & 75.9 & 71.6 & 69.8 & 70.3 & 67.6 & 69.8 & 69.3 & 73.9 & 72.9 & 70.7 & 75.4 & 71.2 & 75.8 & 72.4 & 72.9 & 63.4 & 47.3 & 47.7 & 53.7 & 74.9 & 74.2 & 72.9 & 74.7 & 71.4 \\
    
    \rowcolor{mistralcolor}
    Mistral-Small-24B & 42.8 & 61.2 & 58.2 & 56.5 & 57.7 & 57.8 & 57.7 & 32.0 & 30.8 & 37.6 & 25.7 & 27.2 & 27.8 & 60.7 & 51.8 & 50.3 & 51.1 & 36.1 & 52.6 & 30.2 & 51.8 & 26.4 & 17.1 & 10.2 & 14.8 & 56.8 & 54.9 & 53.1 & 52.4 & 47.9 \\
    \rowcolor{mistralcolor}
    Mistral-Small-3.1-24B & 45.9 & 62.0 & 60.6 & 58.5 & 59.4 & 60.0 & 59.6 & 40.8 & 32.0 & 43.3 & 29.3 & 30.7 & 32.9 & 56.5 & 54.4 & 52.3 & 53.4 & 35.4 & 55.5 & 49.8 & 53.3 & 31.4 & 17.0 & 13.5 & 17.0 & 59.2 & 56.0 & 53.9 & 55.1 & 48.7 \\
    
    \rowcolor{interncolor}
    InternLM3-8B & 17.2 & 40.8 & 38.3 & 36.7 & 36.3 & 36.1 & 34.7 & 5.2 & 3.8 & 2.5 & 6.4 & 2.0 & 1.5 & 24.2 & 20.6 & 20.0 & 5.3 & 5.5 & 31.6 & 9.1 & 27.6 & 2.2 & 0.6 & 0.6 & 2.2 & 26.1 & 27.5 & 28.9 & 0.0 & 22.2 \\
    
    \rowcolor{ayacolor}
    Aya-23-8B & 13.7 & 34.4 & 18.4 & 20.6 & 19.1 & 11.4 & 18.0 & 11.8 & 3.4 & 4.7 & 3.4 & 4.2 & 4.7 & 27.5 & 18.0 & 20.5 & 16.1 & 5.5 & 11.1 & 19.6 & 19.4 & 5.3 & 3.0 & 4.1 & 3.2 & 19.2 & 17.6 & 14.4 & 19.0 & 18.8 \\
    \rowcolor{ayacolor}
    Aya-23-32B & 25.6 & 40.8 & 36.5 & 36.7 & 35.4 & 30.1 & 34.5 & 27.7 & 14.0 & 20.9 & 7.2 & 13.5 & 14.5 & 37.4 & 29.9 & 34.4 & 30.9 & 14.9 & 23.1 & 36.6 & 29.7 & 9.0 & 1.5 & 3.9 & 14.5 & 36.7 & 35.9 & 27.4 & 34.5 & 29.1 \\
    
    \rowcolor{phicolor}
    Phi4-mini-3.8B & 21.6 & 51.6 & 37.3 & 36.8 & 37.9 & 37.7 & 35.3 & 13.0 & 11.7 & 14.0 & 14.9 & 12.7 & 8.0 & 35.3 & 23.9 & 14.3 & 24.6 & 17.6 & 14.0 & 27.4 & 18.0 & 13.7 & 3.3 & 3.4 & 2.1 & 32.6 & 22.8 & 24.4 & 21.1 & 16.8 \\
    \rowcolor{phicolor}
    Phi4-14B & 49.9 & 71.5 & 61.9 & 64.1 & 59.6 & 61.7 & 60.2 & 47.8 & 34.1 & 41.8 & 24.1 & 43.2 & 36.0 & 62.3 & 56.5 & 58.2 & 57.1 & 51.7 & 63.9 & 56.8 & 57.8 & 35.2 & 8.1 & 23.1 & 11.5 & 65.2 & 61.3 & 50.7 & 63.2 & 59.4 \\

    \rowcolor{llamacolor}
    Llama3.1-8B & 22.9 & 45.2 & 31.2 & 32.2 & 20.7 & 37.8 & 34.8 & 20.5 & 18.8 & 14.2 & 14.2 & 19.7 & 17.3 & 31.9 & 24.9 & 24.0 & 31.3 & 27.9 & 22.6 & 13.4 & 27.9 & 15.1 & 0.4 & 6.4 & 4.5 & 28.8 & 24.1 & 27.1 & 23.1 & 25.5 \\
    \rowcolor{llamacolor}
    Llama3.1-70B & 49.3 & 62.1 & 58.2 & 55.4 & 58.3 & 57.3 & 59.2 & 48.9 & 45.7 & 49.1 & 41.8 & 46.0 & 44.6 & 54.6 & 51.7 & 49.5 & 55.5 & 52.8 & 51.3 & 48.6 & 57.1 & 44.6 & 26.6 & 24.3 & 23.8 & 49.2 & 50.3 & 56.1 & 55.9 & 50.5 \\
    \rowcolor{llamacolor}
    Llama3.3-70B & 55.8 & 65.7 & 62.1 & 59.8 & 61.5 & 61.4 & 67.0 & 55.4 & 50.1 & 56.3 & 47.9 & 56.4 & 52.8 & 58.4 & 57.0 & 54.5 & 65.2 & 56.0 & 65.5 & 51.0 & 62.7 & 49.0 & 28.5 & 31.6 & 33.6 & 61.1 & 59.9 & 63.0 & 63.8 & 59.7 \\
    \rowcolor{llamacolor}
    Llama4-Scout & 51.3 & 71.2 & 63.0 & 66.6 & 62.8 & 67.4 & 57.8 & 40.1 & 47.6 & 40.3 & 42.4 & 49.3 & 50.0 & 40.2 & 49.6 & 42.0 & 51.1 & 52.4 & 58.4 & 49.4 & 58.2 & 49.0 & 14.2 & 32.2 & 41.2 & 61.4 & 58.3 & 53.0 & 57.5 & 60.2 \\
            
    \rowcolor{gemmacolor}
    Gemma3-4B & 28.1 & 40.8 & 35.3 & 34.3 & 36.4 & 37.1 & 36.4 & 28.6 & 27.1 & 18.1 & 25.4 & 24.9 & 22.2 & 33.4 & 31.4 & 28.3 & 32.4 & 26.9 & 33.5 & 29.7 & 28.6 & 18.0 & 7.0 & 12.7 & 9.8 & 34.7 & 32.9 & 32.1 & 33.5 & 24.9 \\
    \rowcolor{gemmacolor}
    Gemma3-12B & 48.5 & 59.8 & 55.6 & 54.3 & 55.8 & 55.5 & 56.3 & 49.6 & 46.0 & 43.6 & 46.7 & 47.7 & 47.2 & 51.7 & 50.8 & 49.1 & 54.5 & 49.7 & 55.4 & 49.9 & 50.7 & 44.0 & 9.4 & 25.0 & 32.0 & 54.0 & 53.9 & 53.1 & 54.8 & 50.5 \\
    \rowcolor{gemmacolor}
    Gemma3-27B & 56.6 & 66.5 & 63.5 & 61.0 & 63.0 & 63.2 & 64.4 & 58.4 & 55.5 & 56.7 & 55.9 & 56.1 & 56.8 & 60.4 & 59.3 & 57.8 & 61.1 & 56.7 & 62.6 & 58.7 & 62.0 & 52.8 & 8.8 & 32.4 & 40.7 & 62.5 & 61.7 & 61.7 & 62.6 & 59.8 \\
    
    \rowcolor{qwencolor}
    Qwen2.5-3B & 24.8 & 43.9 & 35.6 & 34.3 & 34.7 & 31.8 & 32.7 & 21.3 & 14.2 & 24.5 & 19.4 & 23.3 & 9.2 & 34.1 & 30.1 & 28.4 & 29.9 & 27.4 & 31.6 & 28.6 & 28.9 & 6.6 & 1.1 & 10.6 & 5.4 & 26.2 & 28.5 & 23.8 & 28.8 & 24.3 \\
    \rowcolor{qwencolor}
    Qwen2.5-7B & 37.6 & 57.5 & 48.9 & 46.9 & 49.3 & 46.1 & 49.1 & 34.0 & 32.2 & 25.7 & 23.4 & 29.5 & 27.3 & 50.5 & 43.6 & 41.5 & 46.4 & 39.6 & 46.6 & 40.2 & 41.1 & 23.2 & 11.0 & 21.1 & 13.3 & 46.3 & 42.9 & 39.7 & 42.3 & 31.3 \\
    \rowcolor{qwencolor}
    Qwen2.5-14B & 47.7 & 65.5 & 59.1 & 58.5 & 59.6 & 59.0 & 59.8 & 43.4 & 41.5 & 41.1 & 33.6 & 39.2 & 38.3 & 59.9 & 54.7 & 53.5 & 57.1 & 50.3 & 55.9 & 51.0 & 53.3 & 24.2 & 26.3 & 25.9 & 8.6 & 59.1 & 54.0 & 50.1 & 53.7 & 46.0 \\
    \rowcolor{qwencolor}
    Qwen2.5-32B & 54.2 & 68.7 & 65.2 & 63.4 & 65.1 & 64.7 & 65.9 & 49.1 & 48.1 & 50.7 & 38.9 & 49.0 & 48.7 & 62.6 & 61.7 & 59.1 & 62.4 & 56.8 & 64.2 & 58.5 & 59.8 & 35.0 & 27.2 & 27.3 & 19.4 & 64.0 & 61.6 & 59.8 & 60.4 & 56.1 \\
    \rowcolor{qwencolor}
    Qwen2.5-72B & 58.8 & 70.3 & 67.1 & 65.9 & 66.5 & 66.6 & 68.8 & 58.0 & 57.6 & 59.1 & 47.9 & 56.4 & 57.3 & 65.9 & 63.4 & 62.1 & 66.4 & 60.1 & 68.2 & 62.1 & 65.4 & 40.1 & 29.0 & 27.9 & 25.9 & 68.1 & 66.1 & 65.4 & 66.8 & 62.3 \\
    \rowcolor{qwencolor}
    Qwen3-4B & 42.0 & 59.2 & 54.5 & 52.6 & 54.6 & 54.3 & 55.1 & 39.3 & 35.5 & 34.3 & 24.4 & 32.6 & 35.7 & 54.3 & 49.7 & 46.7 & 52.5 & 47.1 & 52.9 & 44.5 & 48.7 & 16.2 & 5.6 & 8.9 & 9.0 & 53.6 & 51.2 & 49.1 & 49.8 & 45.8 \\
    \rowcolor{qwencolor}
    Qwen3-4B-Think & 53.8 & 65.2 & 62.9 & 62.9 & 63.3 & 63.4 & 63.1 & 56.9 & 53.8 & 48.9 & 52.3 & 55.3 & 55.1 & 54.3 & 61.3 & 58.7 & 62.3 & 58.9 & 62.6 & 58.0 & 59.7 & 30.4 & 19.3 & 16.7 & 17.3 & 53.8 & 61.2 & 60.2 & 61.6 & 60.0 \\
    \rowcolor{qwencolor}
    Qwen3-8B & 46.0 & 63.3 & 59.8 & 57.8 & 59.8 & 58.7 & 59.7 & 47.0 & 44.3 & 44.6 & 21.1 & 37.2 & 43.3 & 58.5 & 54.8 & 52.9 & 57.3 & 52.3 & 58.9 & 52.1 & 54.6 & 6.9 & 1.3 & 8.4 & 1.6 & 59.1 & 55.6 & 54.7 & 56.4 & 51.6 \\
    \rowcolor{qwencolor}
    Qwen3-8B-Think & 58.8 & 70.6 & 68.3 & 67.9 & 68.4 & 68.0 & 68.6 & 63.2 & 61.3 & 60.7 & 56.8 & 63.0 & 60.4 & 60.0 & 66.7 & 63.7 & 58.9 & 64.9 & 68.1 & 64.4 & 66.7 & 31.5 & 19.9 & 24.1 & 14.2 & 59.8 & 66.9 & 66.7 & 67.1 & 65.4 \\
    \rowcolor{qwencolor}
    Qwen3-14B & 54.0 & 68.6 & 65.2 & 64.4 & 65.7 & 65.6 & 65.8 & 53.7 & 51.7 & 52.1 & 46.5 & 50.9 & 50.9 & 64.2 & 61.8 & 58.9 & 63.7 & 58.8 & 64.5 & 58.2 & 60.2 & 29.0 & 11.5 & 16.8 & 7.4 & 65.0 & 62.3 & 62.2 & 62.2 & 58.5 \\
    \rowcolor{qwencolor}
    Qwen3-14B-Think & 65.4 & 74.7 & 72.2 & 71.4 & 72.2 & 71.6 & 73.0 & 67.9 & 66.8 & 68.0 & 64.6 & 66.9 & 67.3 & 66.5 & 70.8 & 70.0 & 71.4 & 69.8 & 72.3 & 69.1 & 71.3 & 48.0 & 28.3 & 32.5 & 32.3 & 72.2 & 71.3 & 71.6 & 71.1 & 69.8 \\
    \rowcolor{qwencolor}
    Qwen3-30B-A3B & 57.0 & 70.8 & 67.1 & 65.0 & 66.9 & 66.3 & 67.5 & 57.1 & 53.8 & 54.3 & 50.5 & 53.8 & 54.0 & 66.3 & 63.9 & 60.9 & 65.1 & 60.7 & 65.9 & 60.5 & 62.4 & 29.5 & 25.4 & 20.9 & 23.1 & 66.8 & 64.3 & 64.1 & 64.9 & 61.9 \\
    \rowcolor{qwencolor}
    Qwen3-30B-A3B-Think & 64.8 & 75.7 & 74.0 & 73.4 & 74.0 & 73.8 & 72.0 & 69.3 & 68.6 & 69.2 & 69.4 & 69.8 & 69.0 & 70.5 & 72.1 & 71.4 & 38.9 & 70.3 & 74.0 & 70.6 & 72.4 & 49.6 & 24.9 & 19.5 & 26.2 & 70.6 & 72.8 & 72.7 & 73.6 & 72.2 \\
    \rowcolor{qwencolor}
    QwQ-32B & 63.2 & 75.9 & 75.7 & 74.1 & 76.0 & 75.7 & 75.6 & 71.3 & 70.0 & 59.8 & 58.2 & 59.2 & 68.4 & 66.6 & 66.3 & 65.6 & 72.7 & 72.5 & 70.6 & 62.3 & 73.5 & 45.8 & 8.4 & 19.4 & 21.7 & 74.4 & 70.5 & 64.4 & 73.5 & 65.8 \\
    \rowcolor{qwencolor}
    Qwen3-32B & 59.9 & 71.8 & 68.4 & 67.6 & 68.7 & 69.1 & 69.4 & 61.5 & 57.1 & 62.4 & 51.0 & 58.9 & 59.7 & 67.0 & 62.6 & 65.5 & 68.5 & 56.1 & 68.5 & 64.9 & 65.9 & 46.4 & 26.1 & 25.7 & 17.9 & 68.0 & 68.0 & 67.2 & 67.7 & 65.9 \\
    \rowcolor{qwencolor}
    Qwen3-32B-Think & 66.3 & 74.9 & 72.1 & 71.7 & 72.8 & 72.7 & 73.5 & 70.4 & 66.4 & 70.8 & 70.3 & 70.7 & 70.7 & 68.7 & 70.2 & 71.2 & 72.4 & 70.4 & 73.4 & 70.4 & 72.4 & 56.7 & 26.6 & 18.8 & 35.2 & 69.1 & 73.5 & 72.3 & 72.8 & 71.1 \\
    
    \rowcolor{qwencolor}
    Qwen3-235B & 66.7 & 73.5 & 72.5 & 71.3 & 73.2 & 73.1 & 73.7 & 67.6 & 67.7 & 68.7 & 66.7 & 67.7 & 67.8 & 70.5 & 68.8 & 69.6 & 71.4 & 68.8 & 72.5 & 70.1 & 71.1 & 56.3 & 26.6 & 40.2 & 46.2 & 72.9 & 72.5 & 71.1 & 71.8 & 70.1 \\
    \rowcolor{qwencolor}
    Qwen3-235B-Think & 74.9 & \textbf{80.7} & 80.6 & \textbf{80.4} & \textbf{80.7} & \textbf{80.5} & \textbf{80.9} & \textbf{78.7} & \textbf{77.8} & 76.1 & \textbf{77.9} & \textbf{78.5} & 78.1 & 77.4 & \textbf{77.1} & \textbf{78.3} & 72.6 & 77.1 & 79.9 & \textbf{78.7} & 80.6 & 70.8 & 36.9 & 49.3 & 46.4 & \textbf{77.0} & \textbf{78.8} & 80.2 & \textbf{80.5} & \textbf{79.8} \\

    \bottomrule
    \caption{MMLU-ProX 5-shot Results.} 
    \label{app:full_res1}
    \end{longtable}
\end{sidewaystable} 

\begin{sidewaystable}
    \small %
    \setlength{\tabcolsep}{3pt} 
    \centering
    \begin{longtable}{l|c|cccccc|cccccc|cccccc|cccccc|ccccc}
    \toprule
    Model & \textbf{Avg} & \textbf{EN} & \textbf{FR} & \textbf{DE} & \textbf{ES} & \textbf{PT} & \textbf{IT} & \textbf{HI} & \textbf{BN} & \textbf{UR} & \textbf{TE} & \textbf{MR} & \textbf{NE} & \textbf{ZH} & \textbf{JA} & \textbf{KO} & \textbf{VI} & \textbf{TH} & \textbf{ID} & \textbf{AR} & \textbf{AF} & \textbf{SW} & \textbf{WO} & \textbf{YO} & \textbf{ZU} & \textbf{RU} & \textbf{UK} & \textbf{SR} & \textbf{CS} & \textbf{HU} \\
    \midrule
    
    \rowcolor{gptcolor}
    GPT-4.1 & \textbf{71.1} & \textbf{76.1} & \textbf{74.9} & \textbf{76.1} & 73.8 & \textbf{77.3} & 75.5 & \textbf{74.7} & \textbf{72.9} & \textbf{72.6} & 60.3 & \textbf{74.0} & \textbf{74.2} & \textbf{77.1} & \textbf{73.9} & \textbf{75.1} & \textbf{76.4} & \textbf{74.1} & 67.4 & \textbf{75.1} & \textbf{76.2} & \textbf{71.1} & \textbf{45.0} & \textbf{50.5} & \textbf{65.3} & 50.0 & \textbf{76.1} & \textbf{76.2} & \textbf{75.9} & \textbf{75.4} \\
    
    \rowcolor{mistralcolor}
    Mistral-31-24B & 50.4 & 67.5 & 63.9 & 62.5 & 64.1 & 61.7 & 64.9 & 51.1 & 42.6 & 51.0 & 42.1 & 41.9 & 39.6 & 61.1 & 56.2 & 55.6 & 53.8 & 42.9 & 60.9 & 53.6 & 54.2 & 39.2 & 4.8 & 7.4 & 20.6 & 63.1 & 60.8 & 60.8 & 57.2 & 56.4 \\
    
    \rowcolor{qwencolor}
    Qwen2.5-72B & 58.8 & 72.3 & 68.8 & 67.7 & 68.2 & 68.7 & 69.6 & 59.9 & 57.1 & 59.7 & 37.9 & 55.1 & 57.2 & 66.0 & 64.5 & 62.1 & 67.5 & 63.4 & 68.4 & 64.0 & 65.9 & 41.9 & 19.8 & 24.3 & 24.0 & 68.9 & 66.5 & 66.9 & 67.1 & 60.9 \\
    \rowcolor{qwencolor}
    Qwen3-30B-A3B & 49.1 & 66.6 & 62.6 & 66.1 & 67.4 & 67.0 & 68.3 & 47.0 & 40.9 & 40.6 & 31.8 & 39.0 & 35.0 & 65.0 & 63.5 & 62.4 & 63.6 & 45.5 & 48.4 & 59.8 & 55.4 & 21.3 & 0.1 & 0.6 & 8.3 & 59.9 & 52.8 & 62.3 & 64.6 & 58.1 \\
    \rowcolor{qwencolor}
    Qwen3-30B-A3B-Think & 65.0 & 75.5 & \textbf{74.9} & 74.1 & \textbf{75.0} & 74.6 & \textbf{75.8} & 71.2 & 70.4 & 71.4 & \textbf{69.4} & 71.2 & 66.2 & 69.6 & 73.6 & 73.0 & 73.8 & 71.6 & \textbf{75.0} & 71.7 & 73.9 & 55.8 & 0.7 & 0.5 & 13.3 & \textbf{69.5} & 73.8 & 74.0 & 74.8 & 71.0 \\
    \rowcolor{qwencolor}
    Qwen3-32B & 51.3 & 64.2 & 49.5 & 53.8 & 69.5 & 67.9 & 68.7 & 56.3 & 44.3 & 52.2 & 29.0 & 52.9 & 53.4 & 66.0 & 58.8 & 66.1 & 67.3 & 44.8 & 57.7 & 63.9 & 57.2 & 27.4 & 2.1 & 8.3 & 23.3 & 24.8 & 58.1 & 68.2 & 69.1 & 63.9 \\
    \rowcolor{qwencolor}
    Qwen3-32B-Think & 61.9 & 71.1 & 70.4 & 69.1 & 71.8 & 71.5 & 71.5 & 67.0 & 65.4 & 67.6 & 56.3 & 68.4 & 63.3 & 67.3 & 68.5 & 69.9 & 69.7 & 66.6 & 70.9 & 68.0 & 69.5 & 57.2 & 0.8 & 10.0 & 19.1 & 63.6 & 70.8 & 71.2 & 72.1 & 65.2 \\
    
    \bottomrule
    \caption{MMLU-ProX zero-shot Results.} 
    \label{app:full_res2}
    \end{longtable}
    \begin{longtable}{l|c|cccccc|cccccc|cccccc|cccccc|ccccc}
    \toprule
    Model & \textbf{Avg} & \textbf{EN} & \textbf{FR} & \textbf{DE} & \textbf{ES} & \textbf{PT} & \textbf{IT} & \textbf{HI} & \textbf{BN} & \textbf{UR} & \textbf{TE} & \textbf{MR} & \textbf{NE} & \textbf{ZH} & \textbf{JA} & \textbf{KO} & \textbf{VI} & \textbf{TH} & \textbf{ID} & \textbf{AR} & \textbf{AF} & \textbf{SW} & \textbf{WO} & \textbf{YO} & \textbf{ZU} & \textbf{RU} & \textbf{UK} & \textbf{SR} & \textbf{CS} & \textbf{HU} \\
    \midrule

    \rowcolor{gptcolor}
    GPT-4.1 & 71.6 & 78.7 & 75.2 & 73.6 & 77.6 & 74.3 & 79.8 & 73.0 & 70.4 & 66.8 & 65.5 & 69.4 & 72.6 & 73.0 & 73.8 & 74.7 & \textbf{76.4} & 74.3 & 75.9 & 73.0 & 76.9 & 71.1 & 41.5 & 51.7 & 62.9 & 69.9 & 75.9 & 76.0 & 76.2 & 76.5 \\
    \rowcolor{gptcolor}
    o4-mini & 68.0 & 73.3 & 72.4 & 72.4 & 73.1 & 73.1 & 73.6 & 69.6 & 69.7 & 69.7 & 67.7 & 69.0 & 68.4 & 70.1 & 68.2 & 71.6 & 73.5 & 69.6 & 71.4 & 73.0 & 72.3 & 66.8 & 25.0 & 51.9 & 61.7 & 58.8 & 71.9 & 73.5 & 71.9 & 68.4 \\

    \rowcolor{deepseekcolor}
    DeepSeek-R1 & 74.0 & 78.9 & 80.6 & 72.1 & \textbf{80.4} & 77.2 & 78.7 & 75.0 & 65.3 & 74.0 & 70.7 & 68.7 & 76.2 & 75.5 & 75.5 & 73.1 & 75.0 & \textbf{79.3} & 78.6 & 74.8 & 77.2 & \textbf{71.4} & \textbf{59.5} & \textbf{54.3} & \textbf{67.7} & 77.2 & 76.4 & \textbf{79.4} & 75.2 & 78.1 \\
    \rowcolor{deepseekcolor}
    DeepSeek-V3 & 70.1 & 79.6 & 74.7 & 74.8 & 76.7 & 76.4 & 76.9 & 70.2 & 68.0 & 69.2 & 66.2 & 68.4 & 67.5 & 74.5 & 73.0 & 68.5 & 75.2 & 69.7 & 77.7 & 71.9 & 71.8 & 64.8 & 45.2 & 46.8 & 54.6 & 75.9 & 75.2 & 72.1 & 75.0 & 72.8 \\
    
    \rowcolor{phicolor}
    Phi4-14B & 48.5 & 69.0 & 61.1 & 61.1 & 56.8 & 62.6 & 60.5 & 47.6 & 34.4 & 41.8 & 23.3 & 42.9 & 33.8 & 57.8 & 53.9 & 55.6 & 56.1 & 50.9 & 61.1 & 57.5 & 54.4 & 35.9 & 6.5 & 21.8 & 10.9 & 61.7 & 57.1 & 50.9 & 62.8 & 57.7 \\
     
    \rowcolor{gemmacolor}
    Gemma3-27B & 56.0 & 65.6 & 64.8 & 60.5 & 60.4 & 59.9 & 62.8 & 59.4 & 56.1 & 58.2 & 54.6 & 56.8 & 55.8 & 58.7 & 56.6 & 55.6 & 60.9 & 55.8 & 63.6 & 58.0 & 61.6 & 51.9 & 8.7 & 32.5 & 38.4 & 63.6 & 60.5 & 61.9 & 63.8 & 57.1 \\
   
    \rowcolor{llamacolor}
    Llama3.3-70B & 55.7 & 68.0 & 65.6 & 63.6 & 64.5 & 65.1 & 65.0 & 53.9 & 50.7 & 52.7 & 46.6 & 54.9 & 49.5 & 60.0 & 59.0 & 56.3 & 61.2 & 56.0 & 62.4 & 54.8 & 63.4 & 55.1 & 28.1 & 28.4 & 32.1 & 59.2 & 57.7 & 60.0 & 63.3 & 58.3 \\
    \rowcolor{llamacolor}
    Llama4-Scout & 49.8 & 69.0 & 65.1 & 62.9 & 58.5 & 64.8 & 56.3 & 36.7 & 46.6 & 38.6 & 40.1 & 48.8 & 49.1 & 38.6 & 47.8 & 41.3 & 48.0 & 49.3 & 56.6 & 48.8 & 57.0 & 49.0 & 15.0 & 30.8 & 40.1 & 60.7 & 56.3 & 52.9 & 56.3 & 59.0 \\

    \rowcolor{qwencolor}
    Qwen2.5-72B & 59.8 & 70.4 & 69.0 & 68.4 & 69.4 & 68.4 & 68.5 & 59.4 & 59.4 & 61.7 & 49.8 & 56.0 & 57.7 & 67.3 & 65.0 & 65.1 & 66.7 & 62.2 & 66.7 & 62.9 & 64.5 & 41.8 & 28.2 & 28.6 & 25.5 & 70.7 & 67.9 & 63.9 & 66.8 & 61.6 \\
    \rowcolor{qwencolor}
    Qwen3-30B-A3B & 56.0 & 70.4 & 65.8 & 61.6 & 65.3 & 65.6 & 68.0 & 57.0 & 54.6 & 53.1 & 53.4 & 52.4 & 49.1 & 65.3 & 61.9 & 60.5 & 64.1 & 60.0 & 62.4 & 58.3 & 61.1 & 28.7 & 25.7 & 20.2 & 24.7 & 65.3 & 62.1 & 62.4 & 64.1 & 60.9 \\
    \rowcolor{qwencolor}
    Qwen3-30B-A3B-Think & 63.8 & 75.2 & 75.3 & 71.8 & 71.4 & 73.5 & 70.4 & 67.0 & 66.3 & 69.9 & 67.2 & 68.2 & 67.2 & 67.3 & 69.4 & 69.2 & 37.2 & 69.6 & 73.8 & 69.4 & 70.1 & 48.8 & 25.7 & 19.9 & 25.0 & 71.3 & 71.9 & 70.7 & 73.5 & 73.0 \\
    \rowcolor{qwencolor}
    Qwen3-32B & 58.8 & 71.1 & 67.0 & 67.0 & 67.9 & 67.3 & 70.7 & 60.0 & 56.3 & 64.1 & 50.5 & 58.3 & 58.5 & 66.5 & 60.7 & 63.4 & 66.2 & 54.4 & 66.3 & 62.6 & 64.3 & 42.3 & 26.0 & 25.3 & 16.2 & 67.2 & 67.7 & 67.3 & 66.8 & 64.5 \\
    \rowcolor{qwencolor}
    Qwen3-32B-Think & 64.6 & 73.8 & 71.4 & 70.6 & 69.6 & 71.1 & 74.8 & 69.7 & 63.6 & 68.7 & 70.1 & 70.1 & 67.5 & 66.5 & 68.2 & 69.7 & 69.7 & 67.7 & 72.8 & 68.9 & 69.0 & 52.6 & 24.1 & 17.9 & 34.2 & 67.7 & 70.6 & 68.5 & 72.8 & 70.7 \\
    \rowcolor{qwencolor}
    Qwen3-235B & 66.0 & 75.0 & 71.3 & 72.3 & 73.5 & 71.6 & 74.3 & 66.3 & 65.0 & 67.0 & 66.5 & 66.5 & 66.5 & 68.9 & 69.7 & 69.7 & 68.5 & 66.5 & 70.2 & 69.4 & 71.4 & 54.3 & 27.9 & 38.3 & 45.1 & 72.6 & 70.6 & 70.6 & 71.9 & 71.4 \\
    \rowcolor{qwencolor}
    Qwen3-235B-Think & \textbf{74.1} & \textbf{80.4} & \textbf{81.0} & \textbf{78.7} & 79.3 & \textbf{78.4} & \textbf{79.4} & \textbf{76.5} & \textbf{74.8} & \textbf{76.5} & \textbf{75.7} & \textbf{78.9} & \textbf{76.5} & \textbf{76.0} & \textbf{75.9} & \textbf{78.2} & 71.9 & 78.9 & \textbf{81.0} & \textbf{78.1} & \textbf{77.9} & 68.9 & 37.1 & 48.0 & 44.7 & \textbf{78.4} & \textbf{79.1} & 79.3 & \textbf{80.6} & \textbf{79.9} \\
        
    \bottomrule
    \caption{MMLU-ProX Lite 5-shot Results.} 
     \label{app:full_res3}
    \end{longtable}
    \begin{longtable}{l|c|cccccc|cccccc|cccccc|cccccc|ccccc}
    \toprule
    Model & \textbf{Avg} & \textbf{EN} & \textbf{FR} & \textbf{DE} & \textbf{ES} & \textbf{PT} & \textbf{IT} & \textbf{HI} & \textbf{BN} & \textbf{UR} & \textbf{TE} & \textbf{MR} & \textbf{NE} & \textbf{ZH} & \textbf{JA} & \textbf{KO} & \textbf{VI} & \textbf{TH} & \textbf{ID} & \textbf{AR} & \textbf{AF} & \textbf{SW} & \textbf{WO} & \textbf{YO} & \textbf{ZU} & \textbf{RU} & \textbf{UK} & \textbf{SR} & \textbf{CS} & \textbf{HU} \\
    \midrule
    \rowcolor{gptcolor}
    GPT-4.1 & 69.2 & 76.0 & 76.7 & 73.8 & 72.3 & 77.2 & 74.5 & 74.1 & 70.1 & 71.6 & 55.3 & 72.1 & 73.8 & 64.3 & 71.4 & 72.6 & 76.2 & 72.3 & 68.5 & 74.8 & 74.1 & 68.2 & \textbf{44.2} & 45.2 & 60.9 & 47.4 & 74.3 & 73.8 & 74.5 & 76.0 \\
    \rowcolor{gptcolor}
    o4-mini & 67.2 & 74.5 & 73.0 & 70.9 & 72.8 & 71.8 & 74.0 & 69.9 & 70.1 & 69.0 & 68.4 & 69.0 & 70.9 & 72.1 & 68.9 & 69.9 & 71.4 & 67.9 & 73.3 & 73.0 & 71.3 & 66.0 & 17.5 & \textbf{54.6} & 62.8 & 38.1 & 72.3 & 71.4 & 71.9 & 72.3 \\

    \rowcolor{deepseekcolor}
    DeepSeek-R1 & 71.7 & 77.6 & \textbf{80.8} & 70.2 & 70.1 & 72.6 & 77.2 & 74.3 & 69.4 & \textbf{75.5} & 68.7 & 74.7 & 69.2 & \textbf{75.2} & 70.2 & 73.1 & \textbf{78.1} & 74.3 & \textbf{80.6} & \textbf{75.2} & 77.9 & \textbf{71.6} & 33.0 & 53.2 & \textbf{66.8} & 66.0 & 76.2 & 78.1 & 72.1 & 76.7 \\
    \rowcolor{deepseekcolor}
    DeepSeek-V3 & 70.1 & 79.1 & 77.2 & 77.7 & 76.4 & 78.1 & 78.9 & 75.2 & 69.7 & 70.9 & 65.0 & 66.2 & 66.0 & 69.7 & 73.3 & 73.6 & 78.2 & 70.9 & 76.7 & 74.3 & 73.5 & 62.8 & 39.8 & 38.3 & 51.7 & 75.2 & 74.3 & 74.0 & 76.7 & 70.2 \\
       
    \rowcolor{phicolor}
    Phi4-14B & 41.1 & 58.7 & 58.2 & 56.5 & 57.3 & 58.7 & 57.5 & 35.7 & 25.3 & 36.4 & 3.4 & 34.5 & 26.5 & 46.9 & 47.3 & 53.6 & 41.3 & 39.3 & 56.6 & 45.1 & 45.4 & 24.7 & 5.3 & 14.5 & 13.4 & 40.3 & 58.2 & 45.6 & 59.0 & 47.4 \\
        
    \rowcolor{gemmacolor}
    Gemma3-27B & 56.4 & 65.8 & 63.1 & 60.5 & 63.8 & 62.8 & 62.1 & 59.0 & 56.3 & 58.5 & 57.8 & 57.8 & 57.0 & 54.1 & 59.4 & 58.5 & 61.9 & 54.1 & 61.9 & 56.1 & 61.4 & 53.2 & 4.1 & 33.8 & 44.0 & 63.8 & 60.5 & 61.6 & 62.1 & 59.9 \\
    
    \rowcolor{llamacolor}
    Llama3.3-70B & 52.9 & 66.3 & 67.7 & 63.6 & 62.9 & 65.6 & 66.7 & 56.5 & 44.2 & 48.6 & 12.8 & 55.6 & 43.5 & 60.2 & 54.4 & 56.0 & 61.6 & 58.8 & 63.6 & 59.4 & 65.0 & 53.9 & 23.3 & 32.3 & 31.8 & 28.1 & 62.6 & 60.2 & 63.9 & 46.3 \\
    \rowcolor{llamacolor}
    Llama4-Scout & 62.1 & 75.3 & 69.4 & 70.6 & 68.0 & 70.2 & 70.1 & 61.6 & 59.5 & 56.5 & 58.8 & 61.4 & 63.6 & 49.3 & 65.3 & 63.3 & 67.5 & 64.8 & 70.7 & 64.5 & 68.2 & 61.6 & 11.1 & 34.5 & 51.9 & 68.5 & 68.2 & 70.7 & 68.9 & 67.2 \\

    \rowcolor{qwencolor}
    Qwen2.5-72B & 58.9 & 73.1 & 68.0 & 63.9 & 68.4 & 68.7 & 69.0 & 60.0 & 58.0 & 59.5 & 38.6 & 53.9 & 55.8 & 67.3 & 62.8 & 65.3 & 67.7 & 62.1 & 66.8 & 64.8 & 67.5 & 42.0 & 20.1 & 25.7 & 27.2 & 69.9 & 65.8 & 67.0 & 67.7 & 61.2 \\
    \rowcolor{qwencolor}
    Qwen3-30B-A3B & 47.9 & 65.1 & 61.9 & 64.6 & 64.6 & 65.6 & 67.9 & 46.1 & 42.7 & 38.6 & 32.8 & 39.1 & 33.5 & 66.3 & 62.9 & 63.3 & 62.2 & 41.8 & 46.1 & 55.8 & 53.2 & 21.6 & 0.0 & 0.9 & 6.6 & 57.3 & 50.0 & 59.4 & 61.7 & 57.0 \\
    \rowcolor{qwencolor}
    Qwen3-30B-A3B-Think & 64.1 & 76.7 & 73.5 & 72.3 & 73.0 & 72.3 & 75.9 & 69.4 & 69.4 & 70.1 & 68.9 & 70.1 & 64.5 & 66.5 & 70.6 & 71.6 & 73.6 & 70.7 & 73.6 & 71.3 & 72.1 & 55.3 & 0.7 & 0.0 & 13.9 & 71.3 & 71.6 & 72.4 & 74.7 & 71.9 \\
    \rowcolor{qwencolor}
    Qwen3-32B & 50.4 & 62.6 & 46.4 & 51.5 & 69.7 & 67.2 & 65.8 & 54.9 & 42.0 & 51.0 & 27.7 & 51.4 & 54.8 & 65.5 & 60.4 & 64.8 & 67.5 & 41.8 & 53.9 & 62.9 & 54.6 & 28.9 & 2.4 & 8.3 & 24.7 & 26.7 & 57.1 & 66.0 & 68.9 & 61.4 \\
    \rowcolor{qwencolor}
    Qwen3-32B-Think & 60.4 & 70.9 & 69.0 & 66.5 & 69.4 & 67.3 & 71.4 & 65.8 & 63.9 & 66.3 & 53.6 & 63.8 & 62.2 & 65.6 & 67.9 & 70.6 & 65.3 & 63.1 & 70.9 & 67.2 & 66.7 & 56.0 & 0.7 & 7.5 & 20.2 & 65.1 & 70.6 & 68.4 & 69.6 & 65.1 \\
    \rowcolor{qwencolor}
    Qwen3-235B & 62.0 & 72.8 & 71.9 & 70.1 & 72.1 & 69.9 & 73.0 & 68.9 & 58.5 & 68.4 & 48.8 & 67.0 & 66.2 & 66.7 & 54.9 & 71.8 & 68.4 & 50.9 & 63.9 & 71.4 & 72.6 & 52.9 & 12.4 & 32.0 & 31.5 & 73.3 & 61.7 & 70.4 & 70.2 & 66.7 \\
    \rowcolor{qwencolor}
    Qwen3-235B-Think & \textbf{73.4} & \textbf{80.1} & 77.2 & \textbf{78.7} & \textbf{80.4} & \textbf{80.1} & \textbf{83.0} & \textbf{77.2} & \textbf{75.7} & 75.2 & \textbf{77.9} & \textbf{76.0} & \textbf{77.6} & 74.3 & \textbf{75.7} & \textbf{77.0} & 75.3 & \textbf{77.9} & 80.1 & 74.8 & \textbf{78.2} & 70.6 & 27.9 & 50.9 & 32.7 & \textbf{77.4} & \textbf{78.1} & \textbf{81.0} & \textbf{80.3} & \textbf{77.7} \\
        
    \bottomrule
    \caption{MMLU-ProX Lite zero-shot Results.}  
    \label{app:full_res4}
    \end{longtable}
\end{sidewaystable}


\end{document}